\documentclass[twoside,11pt]{article}

\usepackage[accepted]{melba}  
\usepackage{amsmath,amsfonts}

\usepackage{graphicx}
\usepackage{multirow}
\usepackage{multicol}
\usepackage{subcaption}
\usepackage{enumitem}
\usepackage{xcolor}
\usepackage{arydshln}
\usepackage{hyperref}
\usepackage{afterpage}
\usepackage[T1]{fontenc}
\hypersetup{
    colorlinks,
    linkcolor={black},
    citecolor={black},
    urlcolor={black}
}
\melbaheading{2022:013}{https://www.melba-journal.org/papers/2022:013.html}{2020}{1-27}{11/2020}{04/2022}{Tan, Hou, Batten, Qiu, and Kainz}{}{}

\ShortHeadings{Foreign Patch Interpolation}{Tan et al.}
\firstpageno{1}

\title{Detecting Outliers with Foreign Patch Interpolation}

\author{\name Jeremy Tan \email j.tan17@imperial.ac.uk\\
\addr Imperial College London, London, UK \AND  
\name Benjamin Hou \email benjamin.hou11@imperial.ac.uk\\
\addr Imperial College London, London, UK \AND 
\name James Batten \email j.batten@imperial.ac.uk\\
\addr Imperial College London, London, UK \AND 
\name Huaqi Qiu \email huaqi.qiu15@imperial.ac.uk\\
\addr Imperial College London, London, UK \AND 
\name Bernhard Kainz \email b.kainz@imperial.ac.uk\\ 
\addr Imperial College London, London, UK\\ 
Friedrich--Alexander University Erlangen--N\"urnberg, DE
}

\begin{document}

% top matter
\maketitle

% abstract
\begin{abstract}%   <- trailing '%' for backward compatibility of .sty file
In medical imaging, outliers can contain hypo/hyper-intensities, minor deformations, or completely altered anatomy. To detect these irregularities it is helpful to learn the features present in both normal and abnormal images. However this is difficult because of the wide range of possible abnormalities and also the number of ways that normal anatomy can vary naturally. As such, we leverage the natural variations in normal anatomy to create a range of synthetic abnormalities. Specifically, the same patch region is extracted from two independent samples and replaced with an interpolation between both patches. The interpolation factor, patch size, and patch location are randomly sampled from uniform distributions. A wide residual encoder decoder is trained to give a pixel-wise prediction of the patch and its interpolation factor. This encourages the network to learn what features to expect normally and to identify where foreign patterns have been introduced. The estimate of the interpolation factor lends itself nicely to the derivation of an outlier score. Meanwhile the pixel-wise output allows for pixel- and subject- level predictions using the same model. Our code is available at~\url{https://github.com/jemtan/FPI}.
\end{abstract}

% keywords

\section{Introduction}
Outliers in medical data can range from obvious lesions to subtle artifacts. This wide range can make it difficult for a single detection system to identify all irregularities. Moreover, examples of outliers are often not available before testing takes place. This makes it difficult to use conventional classification methods that rely on training data to learn how to recognize test images that come from the same distribution. Without knowing what to look for, this task can be challenging even for human radiologists. For example, when focused on a lung nodule detection task, 83\% of radiologists failed to notice a gorilla superimposed on the image~(\cite{Drew2013Gorilla}). This indicates that human attention can cause even experts to be blind to unexpected stimuli. It is infeasible to have radiologists repeatedly scan for every conceivable irregularity. As such, there may be an opportunity for automated systems to support detection, especially if these tools can offer a complementary view of the data.

Recent works have used neural networks to create high performance image recognition systems. These systems typically learn to recognize different image classes based on features that distinguish them from each other. However, outlier classes are not available during training, so it is not known \textit{a priori} which features will be most relevant. 
%However, many of these approaches learn to recognize a class \textit{in contrast} to other classes. In the case of outlier detection, these conditions are not met because only data from one (normal) class is available. Without a known contrasting class, it is not clear which features are most relevant. 

To circumvent this issue, reconstruction-based methods~(\cite{baur2020scale,zimmerer2019unsupervised,alex2017generative,schlegl2017unsupervised}) aim to learn a complete model of the normal data. Abnormalities are then found by comparing the original image to its reconstruction. A key limitation of this approach is that it directly compares pixel intensities under the assumption that intensity differences will be proportional to abnormality.

Self-supervised methods offer an alternative approach to feature learning. These methods employ data augmentation techniques to create their own labelled samples from unlabelled data. Methods such as geometric transformations~(\cite{golan2018deep}) have been successfully applied to outlier detection and outperform reconstruction-based methods on datasets with high variation such as CIFAR-10~(\cite{krizhevsky2009learning}). These methods can tolerate higher variation in normal data because they learn to identify salient structures in the normal class rather than relying on precise reconstruction of every pixel. However, for most cases in medical imaging, all of the major anatomical structures are present, even in abnormal cases. To detect medically relevant outliers, we aim to develop a method that can tolerate variations of normal anatomy while also being sensitive to fine-grained deviations from normal.

We propose a self-supervised task to train a model to learn where and to what degree a foreign pattern has been introduced. The goal is to encourage the model to learn what features to expect normally, given the context, and to be sensitive to subtle irregularities.

We evaluate this approach on an internal evaluation set with synthetic abnormalities and submitted the technique to the 2020 MICCAI medical out-of-distribution (MOOD) analysis challenge~(\cite{zimmerer2020medical}) where it ranked first in both sample and pixel level tasks. We also evaluate our method's ability to detect real medical anomalies using the DeepLesion dataset~(\cite{yan2018deeplesion}).  
%and thereby assumes that all intensity differences are equal and that they will be proportional to abnormality 
%forfeits many of the attributes of neural networks that make modern classifier so successful  
%A major drawback Directly comparing pixel intensities   

%gaussian process for entire image? inference compares to all previous examples

%attractive reconstruction
%even if the recon is perfect, the degree of abnormality is not necessarily proportional to the image space difference, the same intensity level difference should not result in the same abnormality score. Eg. an intensity difference of 10% in a grey matter region may be less abnormal than the same level of difference within the ventricles. Yet, but image intensity score they produce the same result. Likewise a minor difference in random locations may be less significant than correlated differences along a boundary.
%complementary view 
%
%pixel specific gaussian processes

\section{Related Work}
%There are many ways to approach the outlier detection problem. 
%\TODO: add reference to review article
Out-of-distribution (OOD) detection is a broad topic discussed by many communities~(\cite{pimentel2014review,pang2020deep}). Depending on the context, OOD samples may contain minute defects or completely unrelated content. It is often hard to formally define what constitutes an OOD sample, especially without any reference examples. This makes the task inherently heuristic and each approach must accept some assumptions which will impact its ability to detect different types of outliers. One strategy is to choose assumptions that will generalize as broadly as possible and be sensitive to the types of outliers that are of most interest. Most existing methods detect outliers based on reconstruction error, embedding space distances, or more recently, performance on self-supervised tasks.

Before discussing unsupervised methods, it is important to note that there are many supervised and semi-supervised methods for detecting abnormalities. Supervised methods have achieved expert-level performance in detecting breast cancer~(\cite{wu2019deep}), retinal disease~(\cite{de2018clinically}), pneumonia and other chest abnormalities~(\cite{tang2020automated}). Some of these methods also delineate the boundaries of abnormalities, \emph{e.g.}, brain tumor segmentation~(\cite{menze2014multimodal}). Typically, these supervised methods learn from labelled examples of the target class and are not designed to generalize to other types of abnormalities. Alternatively, there are outlier detection methods that use labelled examples from a subset of anomalies with the goal of detecting broader classes of outliers. One example is outlier exposure~(\cite{hendrycks2019deep}), which trains a multi-class classifier on several classes of normal data and tunes the network to make less confident predictions on a set of OOD training samples that do not belong to any of the normal classes. This tuning can help the model to make less confident predictions on OOD samples, even if they come from a different distribution than the OOD training samples. However, for medical anomalies, which can be very subtle, it is not always possible to obtain a relevant OOD training dataset. Since our proposed method uses only normal samples, we focus on comparing to similar unsupervised methods described below.

Reconstruction-based methods attempt to reproduce images using a model of the normal data. This model may be characterised by the bottleneck of an autoencoder~(\cite{atlason2019unsupervised}) or variational autoencoder (VAE)~(\cite{zimmerer2019unsupervised}) or by the latent space of a generative adversarial network (GAN)~(\cite{schlegl2017unsupervised}). Reconstruction-based methods are especially common in medical imaging applications. They allow for pixel-level localization and offer some level of interpretability through the reconstructed images. Baur et al. provide a comparative study with many variants of reconstruction-based methods using brain MRI data~(\cite{baur2021autoencoders}). Autoencoders are versatile and easy to implement across a wide range of datasets and configurations. For example, different variations have been applied to chest X-ray~(\cite{mao2020abnormality}), mammography~(\cite{wei2018anomaly}), and brain CT~(\cite{pawlowski2018unsupervised}) data. Unconstrained, autoencoders run the risk of reconstructing anomalies along with normal anatomy. As such, many methods use some form of regularization on the latent representation. For instance, \cite{zimmerer2019unsupervised} use a VAE, which maps samples to distributions over the latent space and minimizes the Kullback-Leibler divergence between the approximate posterior and a prior. Alternatively, a discriminator can be used to match the distribution of latent codes to a prior; this type of adversarial autoencoder has also been used in outlier detection~(\cite{chen2018unsupervised}). Another option is to eliminate the bottleneck entirely by using a GAN to learn the distribution of normal data. To reconstruct a query image, a latent code can be optimized to find the best match within the learned distribution~(\cite{schlegl2017unsupervised}) or an encoder can be learned to map images directly into latent codes in a single step~(\cite{schlegl2019f}). There are also restorative methods that replace low likelihood regions in the image with samples from a learned prior~(\cite{you2019unsupervised,marimont2021anomaly}). As such, there are multiple strategies for reconstructing the normal components of the input image. Any errors in the reconstruction are then used to highlight anomalies. However, this means that the abnormality score is proportional to intensity differences in the input space. This neglects some of the key advantages of deep learning. Primarily, it fails to make use of learned mappings that bring raw inputs into representations where semantic differences can be distinguished more easily~(\cite{lecun2015deep}).
%enable inputs to be mapped to a representation 
%the use of neural networks that map inputs to a representation

All of the above methods use whole images, but the reconstruction task can also be simplified to focus on patterns at a smaller scale. Patch-level reconstruction can be effective for detecting pathological textures in mammograms~(\cite{wei2018anomaly}). Decomposing an image into smaller patches can also make it easier to train models, such as GAN's, without down-sampling or losing high-resolution texture information~(\cite{alex2017generative}). Even if a model is trained at the patch level, anomaly scores can be recovered at the pixel level by using overlapping patches during inference~(\cite{alaverdyan2020regularized}). Some of these methods are trained using autoencoder or GAN losses, but exploit components other than the reconstruction error to compute anomaly scores. These can include the discriminator of a GAN~(\cite{alex2017generative}) or the latent representation of an autoencoder~(\cite{alaverdyan2020regularized}). Using the embeddings of an encoder has the potential to facilitate semantic distinctions. However, if the encoder is not trained with an appropriate loss, then the representation may not distinguish relevant samples. For example, a discriminator is trained to separate real and generated samples. This does not necessarily make the representation suitable for separating real healthy samples from real pathological samples.  

Other approaches train encoders using losses that are specifically designed for outlier detection. One example of this learns to map training samples to a compact sphere~(\cite{ruff2018deep}). However, without any examples of outliers in the training data, this latent space may accentuate the wrong features, \emph{i.e.}, variations within the normal data that are class invariant. Some embedding approaches introduce a disjoint set of outlier examples~(\cite{bozorgtabar2020salad}) to overcome this issue. However in this work we focus on methods using only normal data.

Self-supervised methods have recently become a popular approach for unsupervised feature learning, especially variants of contrastive predictive coding (CPC)~(\cite{oord2018representation,henaff2019data}). Self-supervised methods have also been used for outlier detection~(\cite{golan2018deep}), in some cases also combined with CPC~(\cite{tack2020csi}). The main principle underlying many of these methods is to transform the images (\emph{e.g.}, rotation) and train a network to identify the transformation. This will sensitize the network to any features that change consistently with the transformation. For example, the brainstem (in a coronal view) may provide a reliable signal for predicting image rotation. However, if the brainstem structure is missing or occluded, the prediction accuracy may go down, indicating a potential outlier. This approach works well for recognizing key characteristics present in normal data. However, in medical images many pathological outliers may still conform to the same global structure as normal data.

Data augmentation and image synthesis play important roles in several outlier detection methods including our proposed method. In natural image datasets, data augmentation has been used to apply affine transformations, blur or sharpen images, or alter the color, brightness, and contrast of images. Methods such as AutoAugment and RandAugment find the most suitable combination of transformations and achieve state-of-the-art performance on supervised tasks through data augmentation alone~(\cite{cubuk2019autoaugment,cubuk2020randaugment}). For medical imaging applications, elastic deformations and image synthesis can help generate more relevant or realistic augmentations~(\cite{nalepa2019data}). Some methods even model artifacts from the imaging modality used for data acquisition, \emph{e.g.}, the bias field in MRI~(\cite{chen2020realistic}). The data augmentation method that is most closely related to ours is Mixup~(\cite{zhang2017mixup}), which has previously been applied to improve brain tumor segmentation~(\cite{eaton2018improving}). Mixup creates convex combinations of samples and their respective labels. This helps regularize the network to behave linearly in-between classes. It also improves generalization and robustness to adversarial examples. Similarly, CutMix~(\cite{yun2019cutmix}) works by copying a patch from one image and placing it into another image. The labels from both of these images are then mixed (as a convex combination) using a mixing factor equal to the patch area divided by the total image area. 

Both Mixup and CutMix use convex combinations of ground truth labels. However, when there is only one class, which is the case in outlier detection, these convex combinations become meaningless. Self-supervised methods solve this problem by creating new classes through augmentations, \emph{e.g.}, geometric transformations. However, these methods detect outliers through a proxy task, \emph{i.e.}, classifying transformations, instead of directly identifying deviations from normal. This can make it harder to recognize more fine-grained, localized irregularities. Classification-based proxy tasks also lack a direct means of locating abnormalities in the image. In this paper, we show that these elements can be combined in a novel way, using convex combinations to create a new class that represents abnormality. This allows us to train directly on the task of estimating deviation from normal. Meanwhile, our patch-level augmentation setup naturally lends itself to pixel-level localization. 

We provide the full details of our proposed method in the following section. Compared to existing methods, our self-supervised task is designed specifically to improve sensitivity to subtle irregularities. We target these cases because 1) they may be more medically relevant and 2) detecting them may be more useful to radiologists since fine-grained outliers typically require more intense scrutiny, time, and energy to detect.  

\section{Method}
%Since examples of outliers are not available during training, it is difficult to train directly on the desired detection task. As such, 
Most self-supervised methods train a network on a proxy task (\emph{e.g.}, identifying geometric transformations~(\cite{golan2018deep})) and subsequently measure abnormality as \textit{failure} to perform this task. Many of these tasks are helpful for detecting the presence (or absence) of prominent structures that appear in the normal class. But medical images often contain more fine-grained outliers, where most major structures are still intact. As such, we propose a patch-level self-supervision task. 

%The closest approximation to the desired detection task would be to modify patches to appear diseased or abnormal. Unfortunately, the appearance of the abnormalities is not known \textit{a priori} and training on a specific type may limit generalization to unforeseen types. As such, we draw on the natural variations within normal data to create a variety of subtle 
To create a variety of subtle outliers we extract the same patch from two independent subjects and replace the patch with an interpolation between both patches. The operation is shown in Eqn.~\ref{equation:patch_interp} where $A$ and $B$ are independent samples, $i$ refers to individual pixels in a patch $h$, and $\alpha$ is the interpolation factor. Note that $A$, $B$, and $A'$ are full sized images. Pixels outside of the patch remain unchanged and whole images are used as inputs. The patch size, $h_s$, patch center coordinates, $h_c$, and the interpolation factor are all randomly sampled from uniform distributions (Eqn.~\ref{equation:patch_size}-\ref{equation:alpha}). The pixel coordinates of the patch define the region that will be extracted from both samples, $A$ and $B$. For volumetric data, each slice is paired with the corresponding slice from a second subject, based on slice indices. For 2D data or data without a uniform number of slices, images are paired randomly. In both cases, we do not perform any registration preprocessing steps on the data. Instead, we exploit the natural variations and misalignment to create diverse training examples. Patches are square unless truncated by image boundaries or in pixels where $A$ and $B$ have the same value. Patch width ranges between 10\% and 40\% of the image width, $d$.

\begin{equation}
%\begin{split}
A'_{i} = (1-\alpha) A_{i} + \alpha B_{i} \;, \; \forall\; i \in h
%\end{split}
\label{equation:patch_interp}
\end{equation}

\begin{equation}
h_{s} \sim U(0.1\cdot d,0.4\cdot d) 
\label{equation:patch_size}
\end{equation}

\begin{equation}
h_{c} \sim U_{2}(0.1\cdot d,0.9\cdot d) 
\label{equation:patch_center}
\end{equation}

\begin{equation}
\begin{aligned}
\alpha &\sim U(0,1) \; \textrm{for continuous } \alpha \textrm{ or}  \\ 
\alpha &\in \{0,0.25,0.50,0.75,1\} \; \textrm{for discrete } \alpha
\end{aligned}
\label{equation:alpha}
\end{equation}

\begin{figure*}[h]
	\centering
	%brain
	\includegraphics[trim={0cm 0cm 0cm 0cm},clip,width=\linewidth]{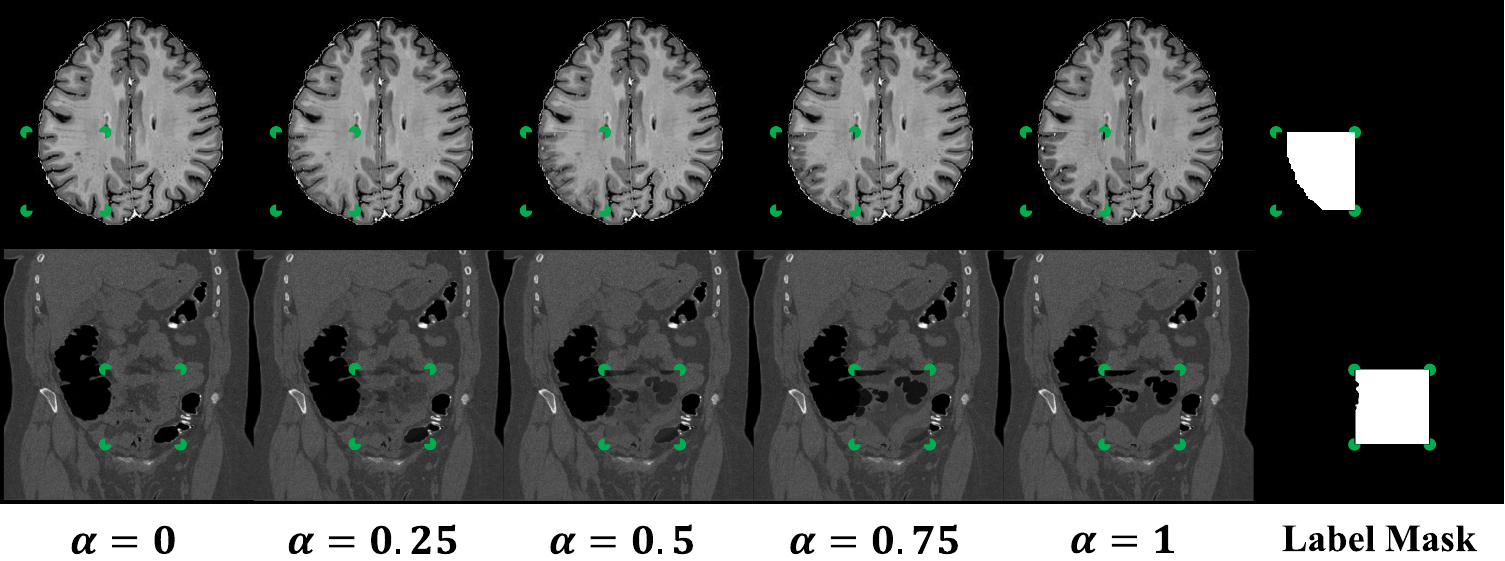}
		%\caption{\textit{Original}}
	
	\caption{\textit{Examples of foreign patch interpolation in brain and abdominal data from the MOOD challenge~(\cite{zimmerer2020medical}). Different $\alpha$ values correspond to different convex combinations. Scaling the label mask by the $\alpha$ value gives the label for each example. Green markers indicate the corners of the patch. Regions where $A$ and $A'$ are equal, \emph{e.g.}, background, are truncated in the label according to Eqn.~\ref{equation:alpha_label}. More examples are given in Appendix~\ref{appendix:fpiBrain} and ~\ref{appendix:fpiAbdomen}.}}
	\label{figures:FPIExamplesShort}		
\end{figure*}

Although $A$ and $B$ are both normal on their own, the differences between them will cause the interpolation, $A'$, to have artificial defects. We train a network to estimate where, and to what degree, a foreign pattern has been introduced. Given $A'$ as input, the corresponding label includes the patch, $h$, and the interpolation factor, $\alpha$, in the form of pixel-level values (Eqn.~\ref{equation:alpha_label}). The loss is thus a pixel-wise regression if $\alpha$ is continuous, or a pixel-wise classification if $\alpha$ is discrete. In both cases a standard cross-entropy loss is used (Eqn.~\ref{equation:loss_regression}-\ref{equation:loss_categorical}, where $f$ represents the model). For continuous $\alpha$, cross-entropy operates on labels that are not one-hot; this is similar to applications such as label smoothing~(\cite{szegedy2016rethinking}), network distillation with soft targets~(\cite{hinton2015distilling}), and MixUp augmentations~(\cite{zhang2017mixup}) and has been studied extensively in its own right~(\cite{NEURIPS2019_f1748d6b,pmlr-v119-lukasik20a}). To obtain predictions during testing, the abnormality score is derived directly from the model's estimate of the interpolation factor $\alpha$. Examples of $A$ and $A'$, with varying alpha, are shown in Figure~\ref{figures:FPIExamplesShort}. The corresponding label for each example is equal to the label mask scaled by the $\alpha$ value.

\begin{equation}
\alpha_i=
\begin{cases}
    \alpha,& \text{if } i\in h \text{ and } A_i\ne B_i\\
    0,              & \text{otherwise}
\end{cases}
\label{equation:alpha_label}
\end{equation}

\begin{equation}
%\mathcal{L}_bce = -y.log(p)-(1-y).log(1-p) 
\mathcal{L}_{\textrm{bce}}(A',\alpha_i,f) = -\alpha_{i}\textrm{log}(f(A'))-(1-\alpha_i)\textrm{log}(1-f(A'))
\label{equation:loss_regression}
\end{equation}

\begin{equation}
\mathcal{L}_{\textrm{cce}}(A',\alpha_i,f) = -\sum_{c=1}^{N=5}\alpha_{i,c}\textrm{log}(f(A'))
\label{equation:loss_categorical}
\end{equation}

Note that FPI does not involve any image registration steps. Nevertheless, it is able to create a range of subtle training samples through simple linear interpolation (as seen in Figure~\ref{figures:FPIExamplesShort} and Appendices~\ref{appendix:fpiBrain} and \ref{appendix:fpiAbdomen}). We experiment on datasets with varying degrees of alignment, \emph{e.g.}, brain MRI volumes with affine registration and CT data with no alignment (details in Section~\ref{sect:eval}). In all cases, FPI is able to form useful training samples that improve detection of outliers.

\subsubsection*{Architecture}
The network architecture is a wide residual encoder-decoder. The encoder portion is a standard wide residual network~(\cite{Zagoruyko2016Wide}) with a width of 4 and a depth of 14. This is designed for inputs with dimensions 256x256. For inputs with dimensions 512x512, an additional residual block is added, bringing the depth up to 16. The decoder follows the same structure as the encoder but in reverse. The terminating activation is sigmoid in the case of continuous $\alpha$ or softmax with the appropriate number of output channels for discrete $\alpha$. 
%add conv group with n = 1, leads to 3 convolutions per group
%but deepest layers have same channel dims so the residual shortcut does not use a conv

\subsubsection*{Training}
Training examples are created dynamically during training with random shuffling of the training data at the start of every epoch. This creates different convex combinations with different samples. 
Each model is trained for 50 epochs using  Adam~(\cite{kingma2014adam}) with a learning rate of $10^{-3}$. An additional training phase can be performed after regular training for stochastic weight averaging~(\cite{izmailov2018averaging}). This step is not necessary to achieve good performance (Figure~\ref{figures:avgPrec}). However, we include its implementation details for completeness. Note that stochastic weight averaging was used in our submission to the MOOD challenge~(\cite{zimmerer2020medical}). To perform stochastic weight averaging, the model is trained for an additional 10 epochs with stochastic gradient descent~(\cite{robbins1951stochastic}) and a cyclic learning rate oscillating in the range $[10^{-4},10^{-3}]$. The varying learning rate helps the model to escape minima and settle in new ones. The parameters are saved whenever the learning rate reaches a minimum (once per epoch). The final model is consolidated by taking the mean of the 10 saved minima. Stochastic weight averaging has been shown to give better generalization~(\cite{izmailov2018averaging}) and approximates ensembling methods without needing to increase model capacity.

\subsection{Evaluation}
\label{sect:eval}
Our method is evaluated on three datasets. The first two come from the MOOD challenge~(\cite{zimmerer2020medical}), while the third is a universal lesion dataset, DeepLesion~(\cite{yan2018deeplesion,yan2018deep}).

\begin{figure*}[h]
	\centering
	%sink/source deform
	\begin{subfigure}[t]{0.31\linewidth}
	    \includegraphics[trim={0.323cm 0.248cm 0.101cm 0.102cm},clip,width=\linewidth,keepaspectratio,angle=90 ]{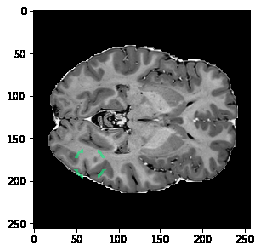}
        \caption{\textit{Original}}
	\end{subfigure}
	\hspace*{\fill}
	\begin{subfigure}[t]{0.31\linewidth}
		\includegraphics[trim={0.323cm 0.248cm 0.101cm 0.102cm},clip,width=\linewidth,keepaspectratio,angle=90]{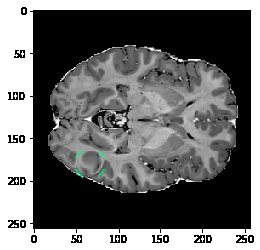}
		\caption{\textit{Synthetic Outlier}}
	\end{subfigure}
	\hspace*{\fill}
	\begin{subfigure}[t]{0.31\linewidth}
		\includegraphics[trim={0.323cm 0.248cm 0.101cm 0.102cm},clip,width=\linewidth,keepaspectratio,angle=90]{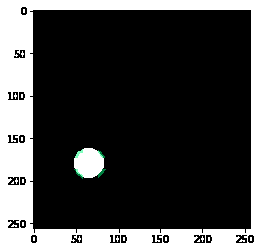}
		\caption{\textit{Label}}
	\end{subfigure}
	\hspace*{\fill}
	
	\caption{\textit{Example of sink/source deformation used to synthesize an outlier. Original sample from MOOD challenge~(\cite{zimmerer2020medical}). All types of synthetic outliers are displayed in Appendix~\ref{appendix:synthetic}.}}
	\label{figures:SyntheticExamplesShort}
\end{figure*}

\noindent\textbf{MOOD Datasets~(\cite{zimmerer2020medical}):} the MOOD challenge provides two datasets, 800 brain MRI volumes (256x256x256) and 550 abdominal CT volumes (512x512x512). Each subject is positioned in approximately the same way, but non-rigid registration is not used. As such, the same voxel/location in two different volumes may contain different tissue. All samples are assumed to be healthy with no abnormalities. Given that no test data is provided, we reserve 10\% of the data as healthy test cases and we use 30\% of the data to create anomalous test cases. The remaining 60\% of the data is used for training. To create the anomalous test set, we synthesize five types of outliers. In each case a sphere of random size and location is selected within each volume; the pixels within that sphere are altered in one of five ways listed below. An example of a sink/source synthetic outlier is given in Figure~\ref{figures:SyntheticExamplesShort}. Performance is evaluated using average precision (AP), which is the metric originally used in the MOOD challenge~(\cite{zimmerer2020medical}). We also include evaluation with area under the receiver operating characteristic curve (AUROC) and an estimated DICE score ($\lceil \textrm{DICE} \rceil$). To compute an approximate DICE score, pixel-level anomaly scores are converted to binary segmentation masks. Following Baur et al., a greedy search is used to find an ideal threshold for this conversion~(\cite{baur2021autoencoders}).
%which is the metric originally used in the MOOD challenge~(\cite{zimmerer2020medical}). We also use 
%Performance is evaluated using receiver operating characteristic (ROC) curves and area under the ROC curve (AUROC).

\begin{itemize}
    \item[$\bullet$] Uniform addition - a sphere of uniform intensity is added to the image;
    \begin{equation}
    A'_i = A_i + n, \; \forall \; i \in h, \text{ where } n \sim \mathcal{N}(0,1)
    \label{equation:synthetic_uniform}
    \end{equation}

    \item[$\bullet$] Noise addition - a sphere of random intensities is added to the image;
    \begin{equation}
    A'_i = A_i + n_i, \; \forall \; i \in h, \text{ where } n_i \sim \mathcal{N}(0,1)
    \label{equation:synthetic_noise}
    \end{equation}
    
    \item[$\bullet$] Sink/source deformation - pixels are shifted toward/away from the center of the sphere;
    \begin{equation}
    \begin{split}
    A'_{I} = A_{V}, \; \forall \; I \in h, \text{ where } I = (i,j,k) \text{ and } \\ V =
    \begin{cases}
    h_c + s(I-h_c),& \text{for source} \\
    I + (1-s)(I-h_c),& \text{for sink}
    \end{cases}\\
    \text{and } s = \left(\frac{\|I-h_c \|_2}{\frac{h_s}{2}}\right)^2
    \end{split}
    \label{equation:synthetic_sinksource}
    \end{equation}

    \item[$\bullet$] Uniform shift - pixels in the sphere are resampled from a copy of the volume which has been shifted by a random distance in a random direction;
    
    \begin{equation}
    \begin{split}
    A'_{i,j,k} = A_{i+a,\; j+b,\; k+c} \; \forall \; i,j,k \in h, \\ \text{ where } a,b,c \sim \sigma\mathcal{U}(0.02\cdot d,0.05\cdot d) \\
    \text{and } \sigma = 
    \begin{cases}
    +1,& \text{with prob. } \frac{1}{2}\\
    -1,& \text{with prob. } \frac{1}{2} 
    \end{cases}
    \end{split}
    \label{equation:synthetic_shift}
    \end{equation}
    
    \item[$\bullet$] Reflection - pixels in the sphere are resampled from a copy of the volume that has been reflected along an axis of symmetry.
    \begin{equation}
    \begin{split}
    A'_{i,j,k} = A_{i,d-j,k} \; \forall \; i,j,k \in h, \\ \text{ where } d \text{ is image width}
    \end{split}
    \label{equation:synthetic_reflect}
    \end{equation}
    
\end{itemize}

\noindent\textbf{DeepLesion Dataset~(\cite{yan2018deeplesion}):} this dataset contains CT scans from 4,427 unique patients exhibiting a broad range of lesions. There are at least eight different types of lesions including lung, abdomen, mediastinum, liver, pelvis, soft tissue, kidney, and bone. Each lesion is annotated with a bounding box. This dataset also includes volumetric data with slices above and below the annotated slice, typically about 30mm on both sides. In many cases, there are multiple annotated slices contained within one volume.
%In many cases, the annotated slice and its surrounding slices are merged with other slices from the same scan to create one continuous volume. 
To extract normal data from these volumes, we remove all annotated slices along with a margin about 10mm on either side. We train on 270,561 normal slices and test on 116,026 normal slices and 4831 annotated slices with lesions. A supervised benchmark is also trained using 22,496 slices with lesions and corresponding bounding box labels. Image-level testing uses normal slices and slices containing lesions. However, for pixel-level evaluation we only use slices with lesions. In this case, pixels inside bounding boxes are considered anomalous and all pixels outside of the bounding boxes are considered normal. All images are resized to 256x256. Performance is evaluated using AUROC and $\lceil \textrm{DICE} \rceil$. Receiver operating characteristic (ROC) curves are also plotted.

The annotations in this dataset are mined from radiology reports and the creators of DeepLesion acknowledge that there may be lesions that have not been annotated, \emph{e.g.}, those that were not relevant to the radiologist's examination~(\cite{yan2018deeplesion}). While this makes training more difficult for outlier detection methods (and also for supervised methods), it represents a more realistic scenario, where it is difficult to ensure that the normal data contains no abnormalities of any kind. %This dataset is also challenging for unsupervised methods because lesions can be very subtle relative to the variance of anatomical structures in the normal data.
%some images are cropped to focus the field of view on the anatomy of interest

\subsection{Benchmark Methods}
To evaluate the performance of the proposed method, foreign patch interpolation (FPI), we compare with several benchmark methods. For the MOOD challenge data with synthetic outliers, we compare with deep support vector data description (SVDD)~(\cite{ruff2018deep}), a convolutional autoencoder (CAE)~(\cite{masci2011stacked}), and a maximum-mean discrepancy VAE (MMD-VAE)~(\cite{zhao2019infovae}). For the DeepLesion data with real medical abnormalities, we compare with several more advanced benchmarks including MMD-VAE~(\cite{zhao2019infovae}), a hierarchical vector-quantized VAE (VQ-VAE2)~(\cite{Razavi2019VQVAE2}), a restoration approach with VQ-VAE2~(\cite{you2019unsupervised,marimont2021anomaly}), and a supervised method.

\noindent \textbf{Deep SVDD}~(\cite{ruff2018deep}) is an embedding-based approach that learns a compact representation of the normal data. The network used is a convolutional encoder with equivalent depth to the encoder of FPI. 

\noindent \textbf{CAE}~(\cite{masci2011stacked}) is a reconstruction-based method. It reconstructs images using features that are learned from normal data. Errors in the reconstruction are then used to highlight abnormal regions. The architecture for the CAE is a convolutional network with equivalent depth to the FPI network.

\noindent \textbf{MMD-VAE}~(\cite{zhao2019infovae}) uses maximum-mean discrepancy (MMD)~(\cite{Gretton2007MMD}) to measure the distance between a prior and the distribution of encodings from real samples. Compared to conventional VAE's, this method is more stable during training and produces high fidelity reconstructions. For our implementation of MMD-VAE, we use the same wide residual encoder-decoder as FPI. Fully connected layers are added to the bottleneck resulting in latent codes of dimension 128. 

\noindent \textbf{VQ-VAE2}~(\cite{Razavi2019VQVAE2}) compresses inputs by quantizing latent codes into discrete values at two levels of the network. We implement VQ-VAE2 using the same wide-residual encoder decoder network as FPI. Vector-quantization is performed at the two deepest layers (closest to the bottleneck). These have dimensions 32x32 and 16x16 respectively. At both levels, latent codes are quantized into 128 discrete values. The activations from the second deepest layer of the encoder are combined with the output of the first layer of the decoder. This skip connection structure allows VQ-VAE2 to produce more accurate reconstructions~(\cite{Razavi2019VQVAE2}).

\noindent \textbf{VQ-VAE2 Restoration}~(\cite{you2019unsupervised,marimont2021anomaly}) uses two PixelCNN models~(\cite{van2016pixel,oord2016conditional}), one at each of the vector quantized layers of the VQ-VAE2. Note that the second PixelCNN takes the latent codes from the first PixelCNN as a conditional input. After learning the distribution of the latent codes, the PixelCNN models can be used to estimate the likelihood of each discrete code. Codes that are deemed to have a low likelihood are discarded and resampled from the learned distribution. The corrected codes are then used to produce a restored image and an anomaly scores is computed from the reconstruction error. Both PixelCNN models are composed of four residual blocks with masked convolutions and four masked convolutional layers on their own.

\noindent \textbf{StyleGAN implementation of AnoGAN}~(\cite{karras2019style,schlegl2017unsupervised}) is a reconstruction-based approach that aims to find a normal version of the query sample in the latent space of a GAN. In this case, a StyleGAN~(\cite{karras2019style}) is used. The model is trained from scratch at progressively higher resolutions, which improves stability and helps to produce more detailed, high resolution images. Instead of using a single latent code as input to the generator, StyleGAN maps a latent code into multiple style codes that are used to control adaptive instance normalization layers throughout the generator~(\cite{huang2017arbitrary}). Gaussian noise is also added at different layers throughout the generator as a source of variation. To reconstruct a query image, we sample 80 initial sets of latent codes and noise vectors and find the set that gives the lowest reconstruction error. Then we further optimize the latent code and noise vectors to minimize the reconstruction error with 20 gradient steps. 

\noindent \textbf{Supervised} training is also done for comparison. Unlike all other benchmarks, which are trained on only normal data, this supervised method is trained on only abnormal data. Lesion bounding boxes are used as labels. The network architecture is the same as the wide residual encoder decoder used in FPI. As such, this benchmark is trained in the same way as FPI, except the labels are real lesion bounding boxes rather than synthetic patch masks. Other more sophisticated supervised methods use region proposal networks to identify and classify patches~(\cite{yan2018deep}). But our arrangement allows us to directly assess the value of ground truth annotations compared with artificial labels generated by FPI.

\section{Results}
We first evaluate FPI on the MOOD challenge data and our synthetic testset. This includes an ablation study and comparison with simple baselines. Then we present results on the DeepLesion dataset and compare with more advanced benchmark methods.

\subsection{MOOD Datasets with Synthetic Anomalies}
\begin{figure*}[h]
	\centering
	\includegraphics[trim={1.29cm 2.16cm 1.47cm 1.93cm},clip,width=0.92\linewidth]{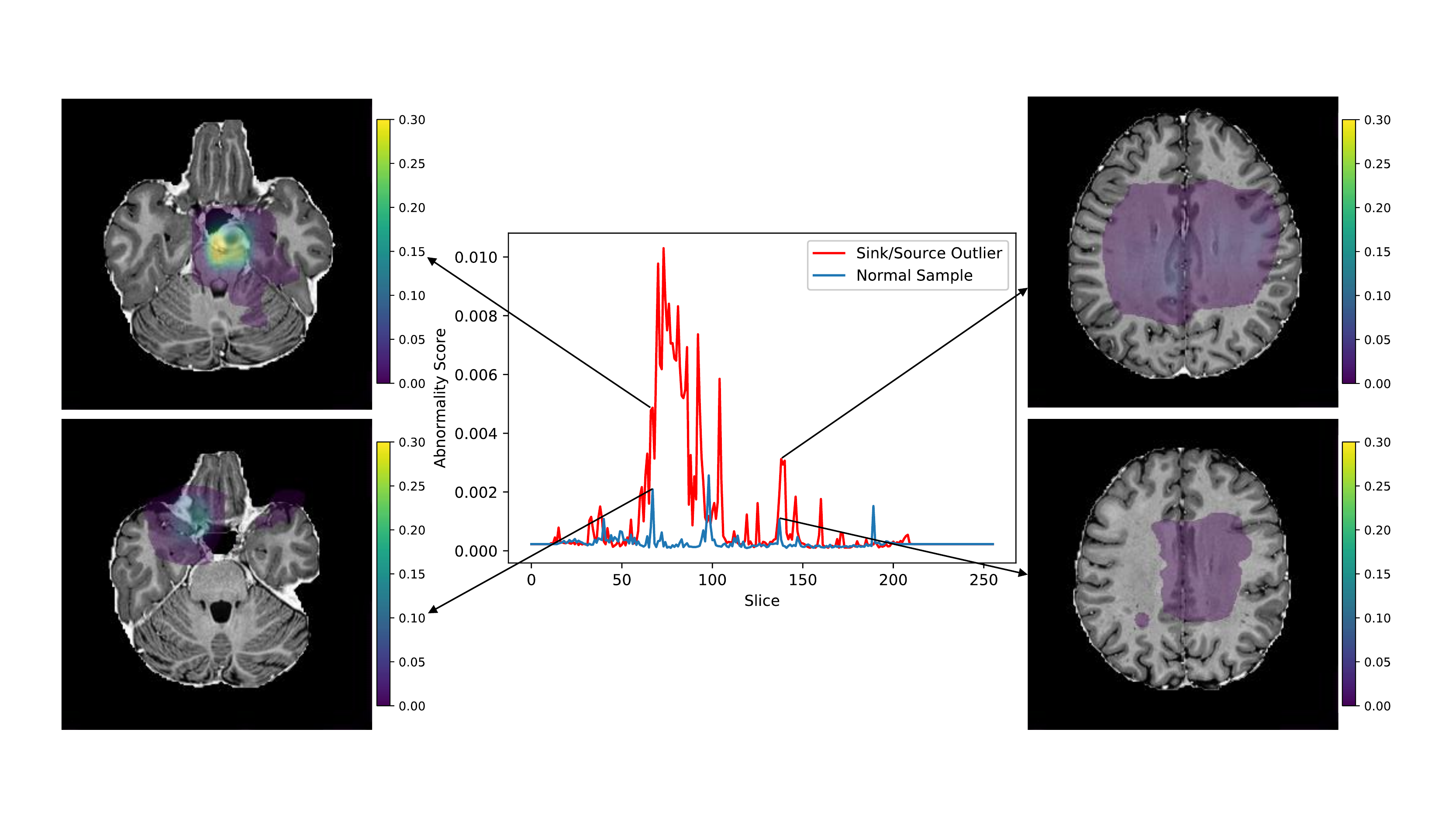}
	\caption{\textit{Image-level abnormality scores for slices throughout the volume. Showing sink/source outlier (red plot and top images) and normal sample (blue line and bottom images). Data from MOOD~(\cite{zimmerer2020medical}). Slices with deformation have high anomaly scores (red), concentrated around the bulbous deformation (top left image). Normal sample has minimal abnormality scores.}}
	\label{figures:sliceAct}		
\end{figure*}

Using the synthetic test data described in Section~\ref{sect:eval}, we evaluate the method's ability to detect different types of outliers. Figure~\ref{figures:sliceAct} displays the model's response to a sink/source deformation outlier and a normal sample. The plot includes abnormality scores for individual slices across the entire volume. Slices that include the artificially deformed sphere produce a strong and consistent activation (Figure~\ref{figures:sliceAct}, red). Meanwhile, normal slices elicit only weak activations (Figure~\ref{figures:sliceAct}, blue).

\begin{figure*}[h]
	\centering
	\includegraphics[clip,width=0.85\linewidth]{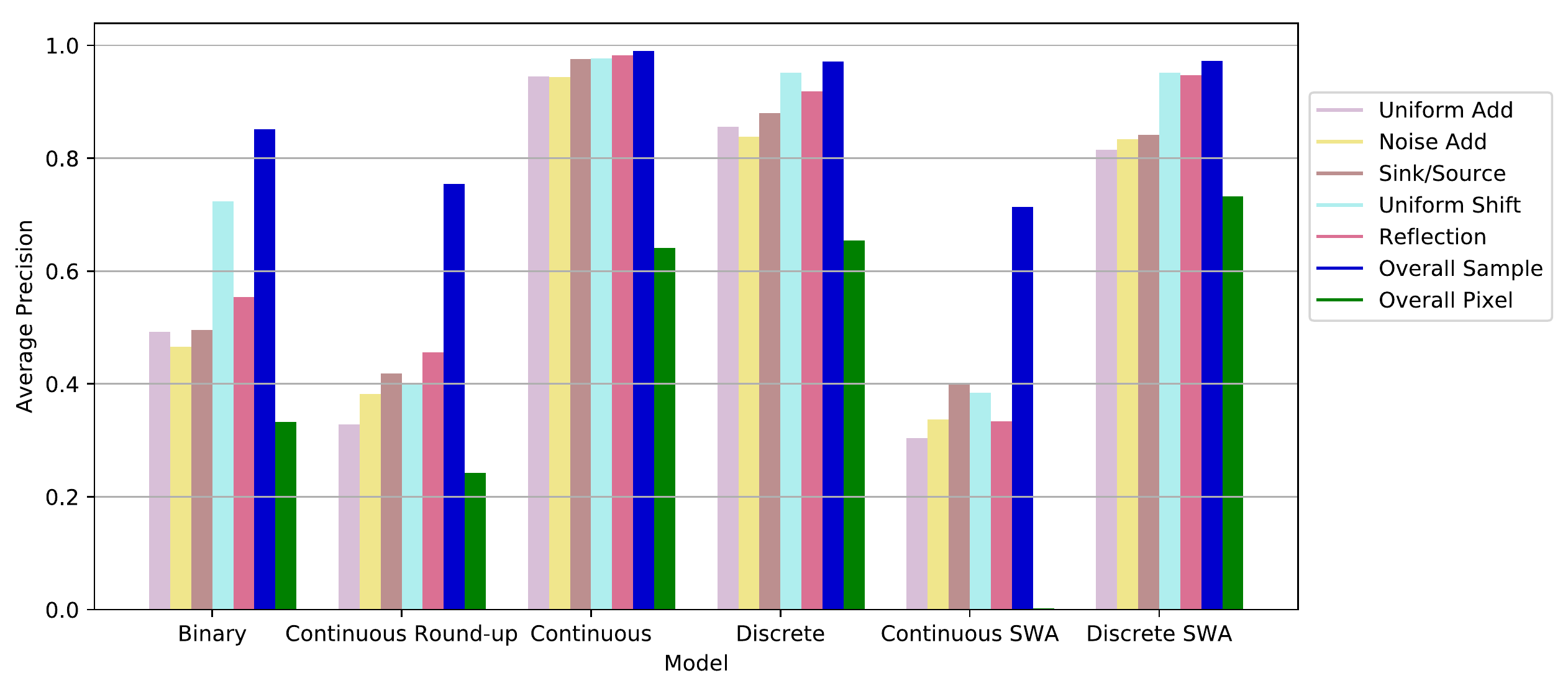}
	\caption{\textit{Average precision for MOOD brain data~(\cite{zimmerer2020medical}) using different model configurations. The binary and continuous round-up models serve as simplified methods used in our ablation study. The continuous and discrete models represent our standard method. The addition of SWA is an optional extension.}}
	\label{figures:avgPrec}		
\end{figure*}

We perform an ablation study by modifying the self-supervision task. A `binary' model is trained using a binary interpolation factor ($\alpha \in \{0,1\}$). For a `continuous round-up' model, the training examples are generated using a continuous interpolation factor ($\alpha \in [0,1]$), but the label supplied to the model is binary ($\alpha = 1 \; \textrm{if} \; \alpha > 0$). We also compare continuous and discrete configurations ($\alpha \in \{0,0.25,0.50,0.75,1\}$) as well as the application of stochastic weight averaging. Figure~\ref{figures:avgPrec} displays the results for individual types of outliers and also overall sample and pixel level scores. Note that the overall scores (Figure~\ref{figures:avgPrec}, blue and green) are calculated using all outlier samples and all normal samples, so the class distribution is different from the individual scores. The binary and continuous round-up models are not able to detect the outliers in the test set effectively. Both continuous and discrete models achieve high performance, even without stochastic weight averaging. The low performance of the continuous stochastic weight averaged model may indicate that optimization is less stable for the continuous task. In contrast, stochastic weight averaging does not hurt performance for the discrete model and can substantially improve pixel-level scores.

The abdominal models were trained in a similar manner and the discrete stochastic weight averaged model achieved the best overall performance. Table~\ref{tab:finalModel} shows the performance of the final selected models which are both trained using the discrete stochastic weight averaged configuration.

\begin{table}[ht]
    \centering
    \caption{Evaluation on synthetic test data, originally from brain and abdominal MOOD data~(\cite{zimmerer2020medical}).}
    \def\arraystretch{1.1}
    \resizebox{\textwidth}{!}{
    \begin{tabular}{c|c|c|c|c|c|c} 
      %\textbf{Anatomy} & \textbf{Method} & 
      %\textbf{Sample-level AP} & \textbf{Pixel-level AP}\\
      \multicolumn{1}{c|}{\multirow{2}{*}{Anatomy}} &
      \multicolumn{1}{c|}{\multirow{2}{*}{Method}} & 
      \multicolumn{2}{c|}{\multirow{1}{*}{Subject-level}} & \multicolumn{3}{c}{\multirow{1}{*}{Pixel-level}}\\
      \multicolumn{1}{c|}{\multirow{1}{*}{}} &
      \multicolumn{1}{c|}{\multirow{1}{*}{}} & 
      \multicolumn{1}{c|}{\multirow{1}{*}{AP}} & \multicolumn{1}{c|}{\multirow{1}{*}{AUROC}} &
      \multicolumn{1}{c|}{\multirow{1}{*}{AP}} & \multicolumn{1}{c|}{\multirow{1}{*}{AUROC}} &
      \multicolumn{1}{c}{\multirow{1}{*}{$\lceil \textrm{DICE} \rceil$}}\\
      \hline
      
      %Brain
      \multicolumn{1}{c|}{\multirow{4}{*}{Brain}} &
      \multicolumn{1}{c|}{\multirow{1}{*}{Deep SVDD~(\cite{ruff2018deep})}} & 0.7695 & 0.5058 & -- & -- & -- \\
      \multicolumn{1}{c|}{\multirow{1}{*}{}} &
      \multicolumn{1}{c|}{\multirow{1}{*}{CAE~(\cite{masci2011stacked})}} & 0.7617 & 0.4947 & 0.0120 & 0.8695 & 0.0269 \\
      \multicolumn{1}{c|}{\multirow{1}{*}{}} &
      \multicolumn{1}{c|}{\multirow{1}{*}{MMD-VAE~(\cite{zhao2019infovae})}} & 0.7572 & 0.4925 & 0.0144 & 0.8790 & 0.0350\\
      \multicolumn{1}{c|}{\multirow{1}{*}{}} &
      \multicolumn{1}{c|}{\multirow{1}{*}{FPI (ours)}} & \textbf{0.9723} & \textbf{0.9321} & \textbf{0.7319} & \textbf{0.9852} & \textbf{0.7092} \\
      \hline
      
      %Abdominal
      \multicolumn{1}{c|}{\multirow{4}{*}{Abdomen}} &
      \multicolumn{1}{c|}{\multirow{1}{*}{Deep SVDD~(\cite{ruff2018deep})}} & 0.8318 & 0.5648 & -- & -- & -- \\
      \multicolumn{1}{c|}{\multirow{1}{*}{}} &
      \multicolumn{1}{c|}{\multirow{1}{*}{CAE~(\cite{masci2011stacked})}} & 0.7378 & 0.4717 & 0.0096 & 0.7240 & 0.0285 \\
      \multicolumn{1}{c|}{\multirow{1}{*}{}} &
      \multicolumn{1}{c|}{\multirow{1}{*}{MMD-VAE~(\cite{zhao2019infovae})}} & 0.7356 & 0.4737 & 0.0079 & 0.7228 & 0.0235 \\
      \multicolumn{1}{c|}{\multirow{1}{*}{}} &
      \multicolumn{1}{c|}{\multirow{1}{*}{FPI (ours)}} & \textbf{0.8854} & \textbf{0.8025} & \textbf{0.6229} & \textbf{0.9292} & \textbf{0.6354} \\
      
    \end{tabular}%
    }
    \label{tab:finalModel}
\end{table}%

Since FPI is trained on synthetic examples, \emph{i.e.}, interpolated patches, it is able to detect other similar classes of synthetic anomalies relatively easily. In comparison, reconstruction-based methods have difficulty identifying these synthetic anomalies because they have minimal intensity differences and occupy less than 1\% of the total imaging volume of a subject. Although the reconstruction-based methods have high scores for pixel-level AUROC, the DICE scores are quite low (Table~\ref{tab:finalModel}). This is because the DICE score focuses more on anomalous pixels, while AUROC can be partly inflated by a large number of normal background pixels. These blank pixels are easy to reconstruct without error. This increases the number of true negatives, which in turn decreases the false positive rate (x-axis of the ROC curve) and increases the area under the curve. In contrast, pixels in tissue regions often have some level of reconstruction error because there is a limit to the amount of detail that the models can recreate. Since the synthetic anomalies have similar intensity values, they also produce similar reconstruction error. When the reconstruction error is averaged across the entire volume, the contribution from the synthetic anomaly is hidden by the contributions from other healthy regions, which leads to a poor subject-level AUROC. Meanwhile, FPI produces very low anomaly scores for normal tissue and activates specifically for certain types of features, as seen in Figure~\ref{figures:sliceAct}.

\begin{figure*}[h]
	\centering
	\begin{subfigure}[b]{0.5515\linewidth}%0.647,0.5805
        \includegraphics[clip,width=\linewidth]{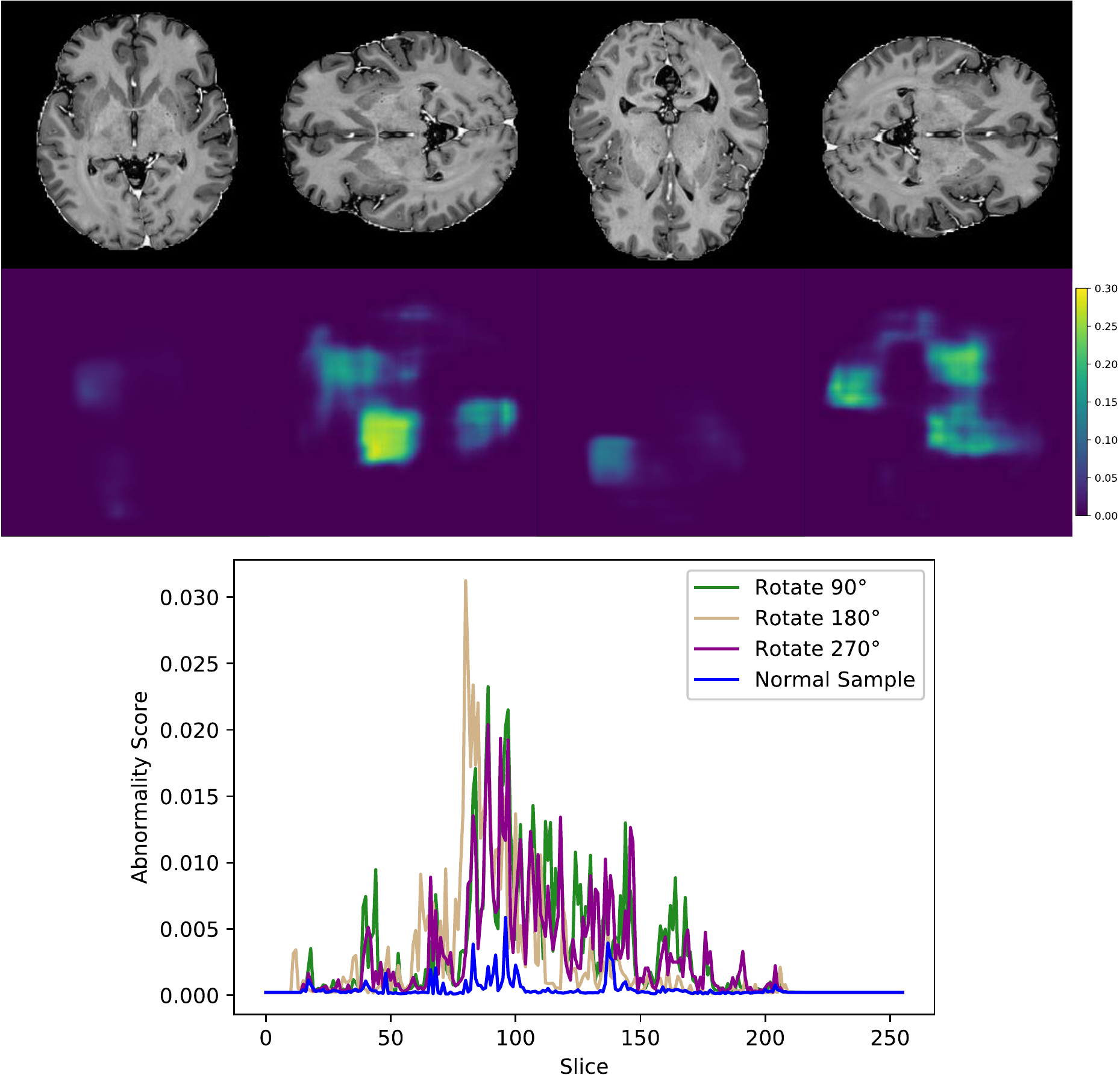}
		\caption{\textit{Rotations}}
	\end{subfigure}
	\hspace*{\fill}
	\begin{subfigure}[b]{0.3696\linewidth}%0.3315,0.389
        %0.342cm 0.314cm 20.168cm 0.243cm
        %trim={20.503cm 0.787cm 0.751cm 0.081cm}
		\includegraphics[clip,width=\linewidth]{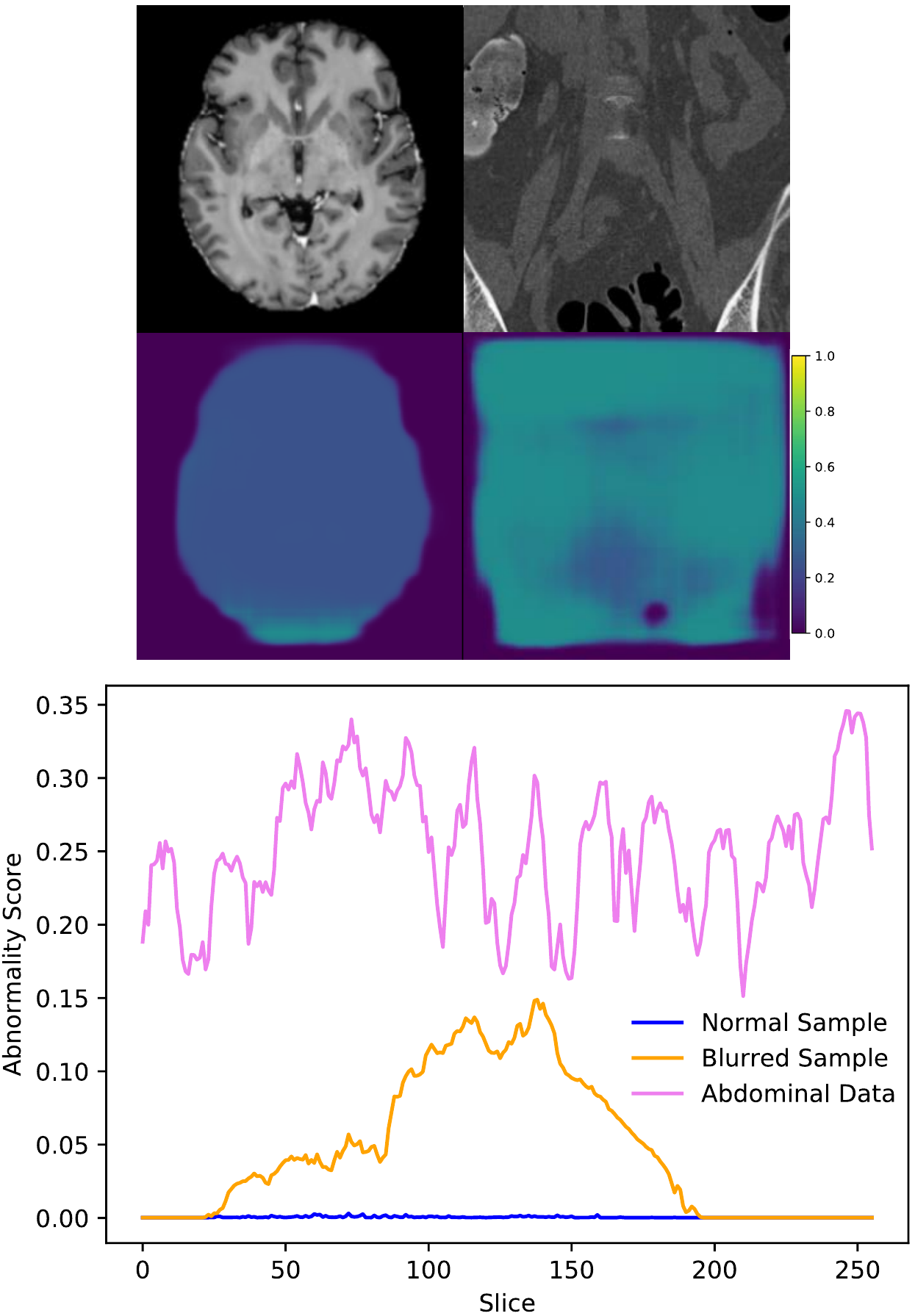}
		\caption{\textit{Blurred and abdominal data}}
	\end{subfigure}
	\hspace*{\fill}
	\caption{\textit{Examples of global outliers using MOOD data~(\cite{zimmerer2020medical}). (a) Original normal sample (top left) and rotations. (b) Gaussian blur ($\sigma=1$) and abdominal data. Note the change of scale in activation maps. Plots display the abnormality score across slices.}}
	\label{figures:globalOutliers}		
\end{figure*}
%\clearpage

In addition to the synthetic test set, which only includes local abnormalities, we provide examples of global abnormalities in Figure~\ref{figures:globalOutliers}. A normal sample produces minimal activation in its canonical orientation (Figure~\ref{figures:globalOutliers}, left most image in (a)). However, rotating the sample produces scattered activations throughout the entire volume (Figure~\ref{figures:globalOutliers}, (a)). Blurring or substituting different anatomy produces even stronger activations (Figure~\ref{figures:globalOutliers}, (b)). 
%\FloatBarrier
%\clearpage
\subsection{DeepLesion Dataset with Medical Anomalies}
\begin{figure*}[h]
	\centering
	\begin{subfigure}[h]{0.4\linewidth}
        \includegraphics[clip,width=\linewidth]{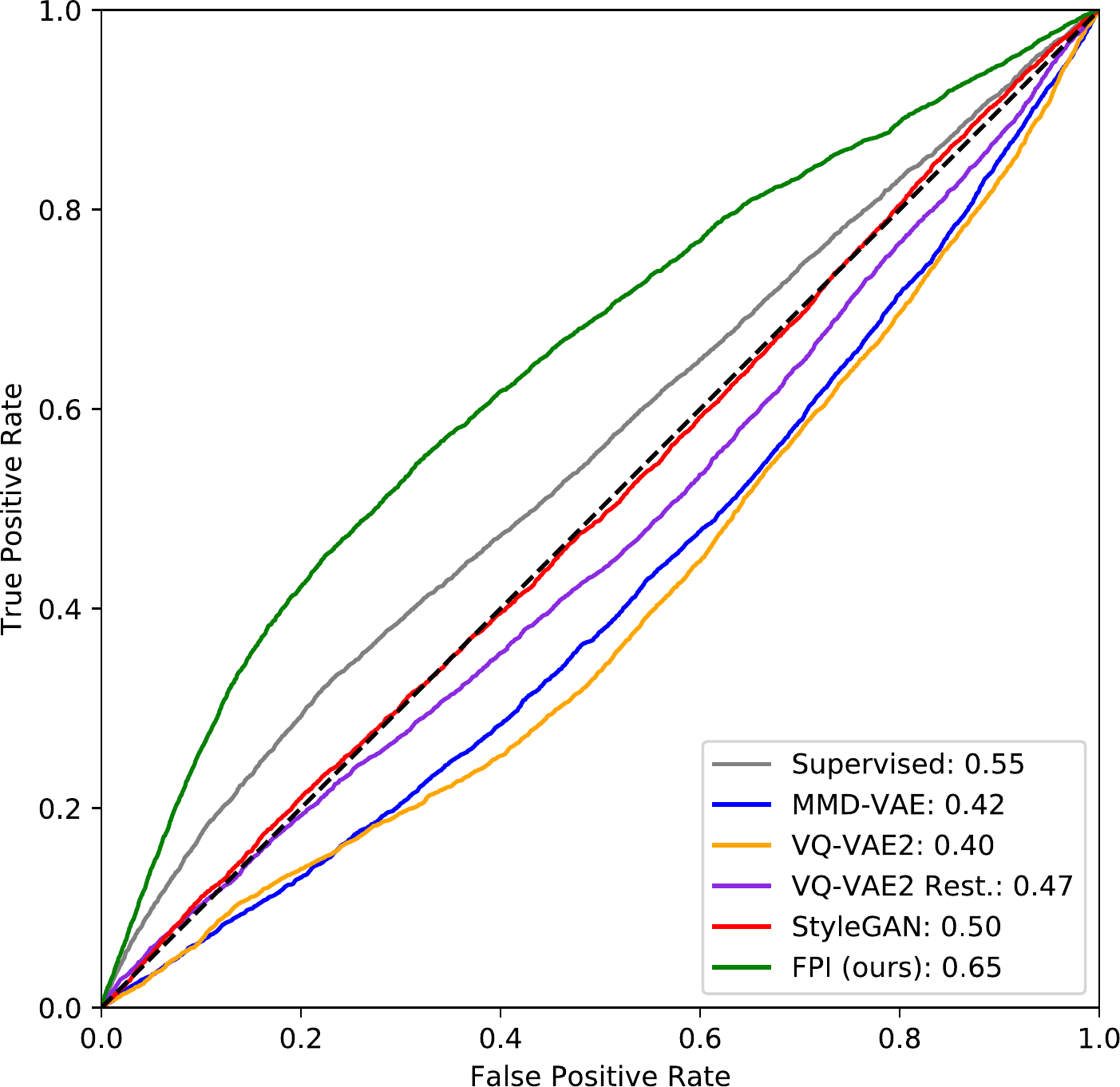}
		\caption{\textit{Image-level}}
		%0.48
	\end{subfigure}
	\hspace*{\fill}
	\begin{subfigure}[h]{0.4\linewidth}
        \includegraphics[clip,width=\linewidth]{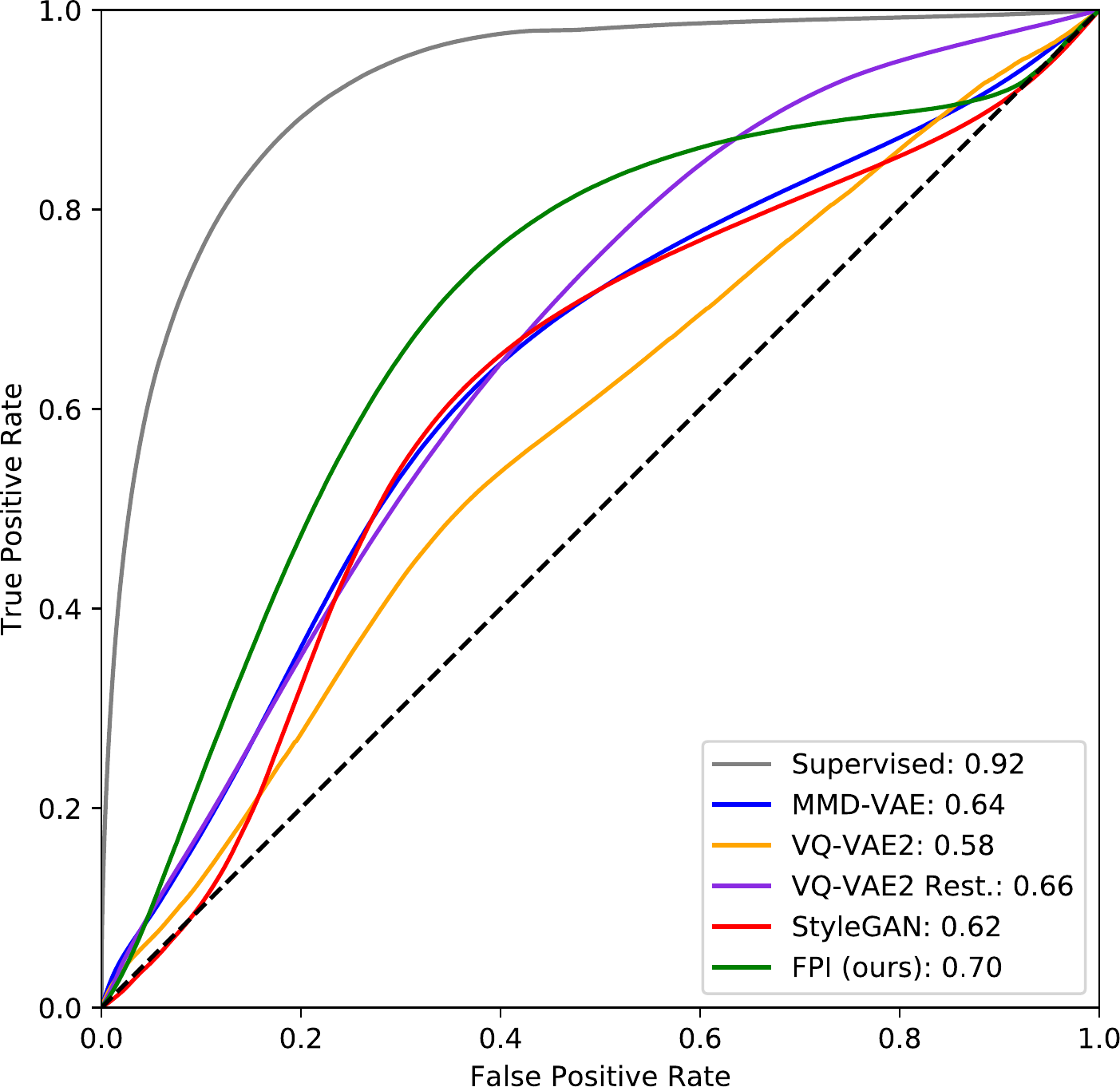}
		\caption{\textit{Pixel-level}}
	\end{subfigure}
	\hspace*{\fill}
	\caption{\textit{ROC curves for DeepLesion data~(\cite{yan2018deeplesion}) for each method at the image-level (a) and pixel-level (b). AUROC reported in the legend.}}
	\label{figures:DeepLesion_ROC_method}
\end{figure*}

For the DeepLesion dataset, FPI was trained under the continuous $\alpha$ (interpolation factor) setting without stochastic weight averaging. The results demonstrate that FPI can identify real medical anomalies despite being trained on only normal images. Table~\ref{tab:deepLesionAuroc} displays both image and pixel level AUROC scores as well as estimated DICE scores. ROC curves are shown in Figure~\ref{figures:DeepLesion_ROC_method}. 

\begin{table}[ht]
    \centering
    \caption{Evaluation on DeepLesion data~(\cite{yan2018deeplesion}). Image-level evaluation is performed using normal slices and slices with lesions. Pixel-level evaluation is done using only slices with lesions; bounding boxes serve as approximate lesion segmentation masks.}
    \def\arraystretch{1.1}
    \resizebox{\textwidth}{!}{
    \begin{tabular}{c|c|c|c}
      \multicolumn{1}{c|}{\multirow{2}{*}{\textbf{Method}}} & \multicolumn{1}{c|}{\multirow{1}{*}{\textbf{Image-level}}} & \multicolumn{2}{c}{\multirow{1}{*}{\textbf{Pixel-level}}} \\
      \multicolumn{1}{c|}{\multirow{1}{*}{}} & 
      \textbf{AUROC} & \textbf{AUROC} & \textbf{$\lceil \textrm{DICE} \rceil$} \\
      \hline
      \multicolumn{1}{c|}{\multirow{1}{*}{Supervised}} & 0.554 & 0.923 & 0.226 \\
      \hline 
      \multicolumn{1}{c|}{\multirow{1}{*}{MMD-VAE~(\cite{zhao2019infovae})}} & 0.419 & 0.635 & 0.024 \\
      \multicolumn{1}{c|}{\multirow{1}{*}{VQ-VAE2~(\cite{Razavi2019VQVAE2})}} & 0.405 & 0.576 & 0.018 \\
      %\multicolumn{1}{c|}{\multirow{1}{*}{
      \begin{tabular}{c}VQ-VAE2 Restoration \\(\cite{you2019unsupervised,marimont2021anomaly})\end{tabular} 
      & 0.469 & 0.664 & 0.023 \\
      \multicolumn{1}{c|}{\multirow{1}{*}{StyleGAN~(\cite{karras2019style})}} & 0.501 & 0.618 & 0.023 \\
      \multicolumn{1}{c|}{\multirow{1}{*}{FPI (ours)}} & \textbf{0.648} & \textbf{0.701} & \textbf{0.030} \\
      
    \end{tabular}%
    }
    \label{tab:deepLesionAuroc}
\end{table}%

\begin{figure*}[h]
	\centering
    \makebox[\linewidth]{\includegraphics[clip,width=\textwidth]{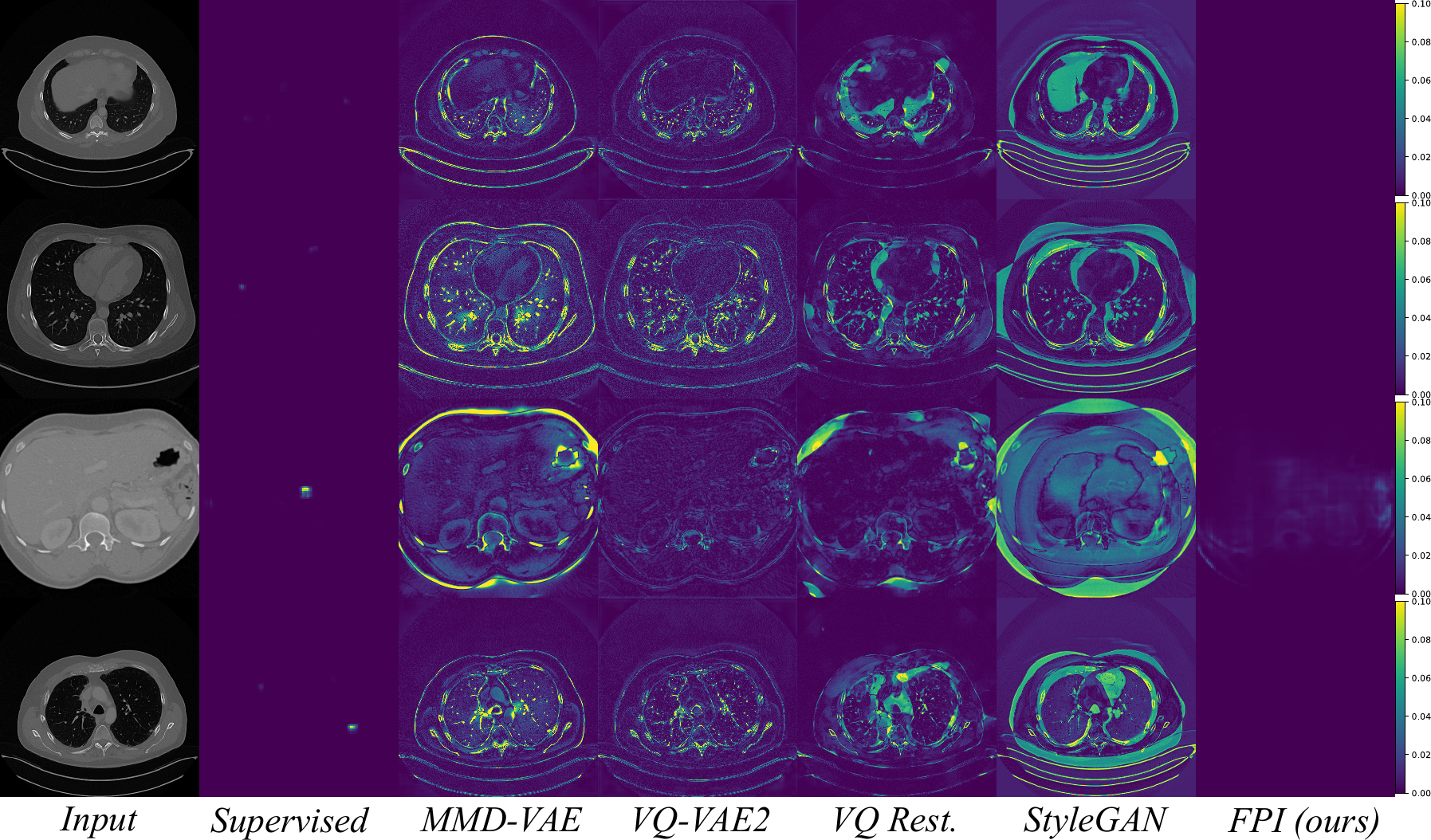}}
    %height=0.78\textheight
	\hspace*{\fill}
	
	\caption{\textit{Normal test samples from DeepLesion~(\cite{yan2018deeplesion}) and outputs from each method. Note that reconstructiont error outputs are scaled down by a factor of five.}}
	\label{figures:DeepLesion_map_norm}
\end{figure*}

At the image level, the reconstruction-based methods score around 0.5 or below. Several factors contribute toward this, including high variation in normal data, higher reconstruction error from certain structures, and overrepresentation of certain tissue types in the normal test data. Figure~\ref{figures:DeepLesion_map_norm} shows that reconstruction-based models must learn to reproduce a wide range of structures and different organs. Most of the reconstruction error comes from sharp edges with high contrast and high spatial frequency, \emph{i.e.}, tissue interfaces. Also, the more pixels involved, the higher the contribution to the overall (image-level) anomaly score. As an example, the lungs generally have a high reconstruction error because they span across a large area and contain details with high spatial frequency. The lungs may also be overrepresented in the normal test data. As described in Section~\ref{sect:eval}, each anomalous test image is accompanied by parallel slices that give context above and below the anomalous slice. The context slices, minus a margin around the anomalous slice, are used as normal test data, resulting in 116,026 normal test images and 4831 anomalous test images. However, certain regions have more context slices than others. For example, the average number of context slices for an anomalous lung image is 79, whereas soft tissue type lesions (muscle, skin, fat) only have 37 context slices on average. As such, the normal test data may be skewed toward certain organs that have high reconstruction error. This can increase the false positive rate and reduce the area under the ROC curve. This skew may exist in the training data as well, but reproducing details with high spatial frequency can still be challenging for methods that rely on a lower dimensional representation of the data. 

The supervised method, which is only trained on slices containing lesions, also performs poorly when tested on images that are both normal and abnormal. This could be fixed by including normal samples during supervised training. But it illustrates that even supervised methods can face difficulty when the test distribution does not match the training distribution. FPI is specifically designed to handle out-of-distribution samples and does not rely on proxy tasks that require full image reconstruction. These properties makes it suitable for detecting subtle lesions within highly variable data.

\begin{figure*}[h]
	\centering
	\begin{subfigure}[h]{0.4\linewidth}
		\includegraphics[clip,width=\linewidth]{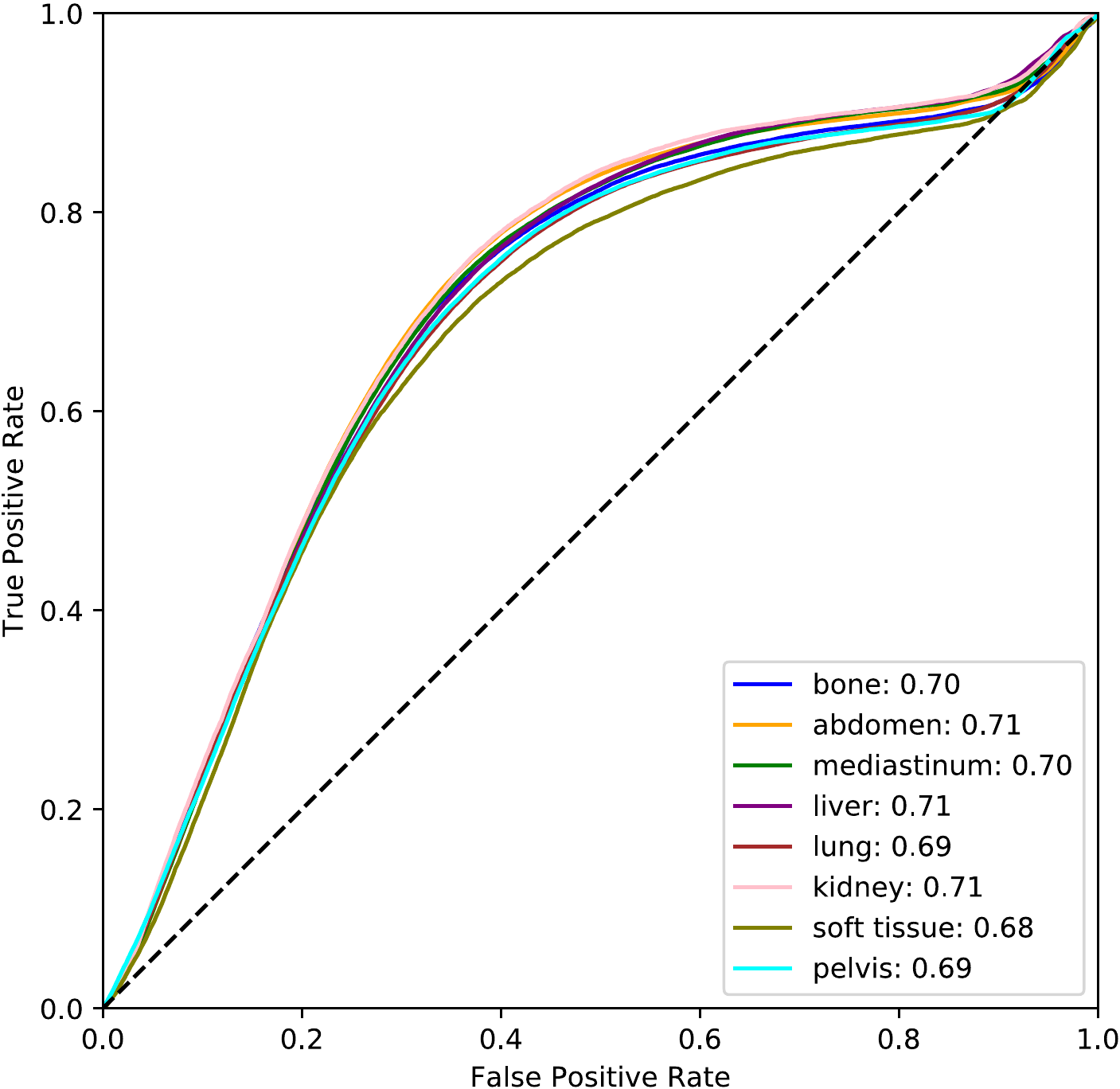}
		\caption{\textit{FPI by lesion type}}
	\end{subfigure}
	\hspace*{\fill}
	\begin{subfigure}[h]{0.4\linewidth}
		\includegraphics[clip,width=\linewidth]{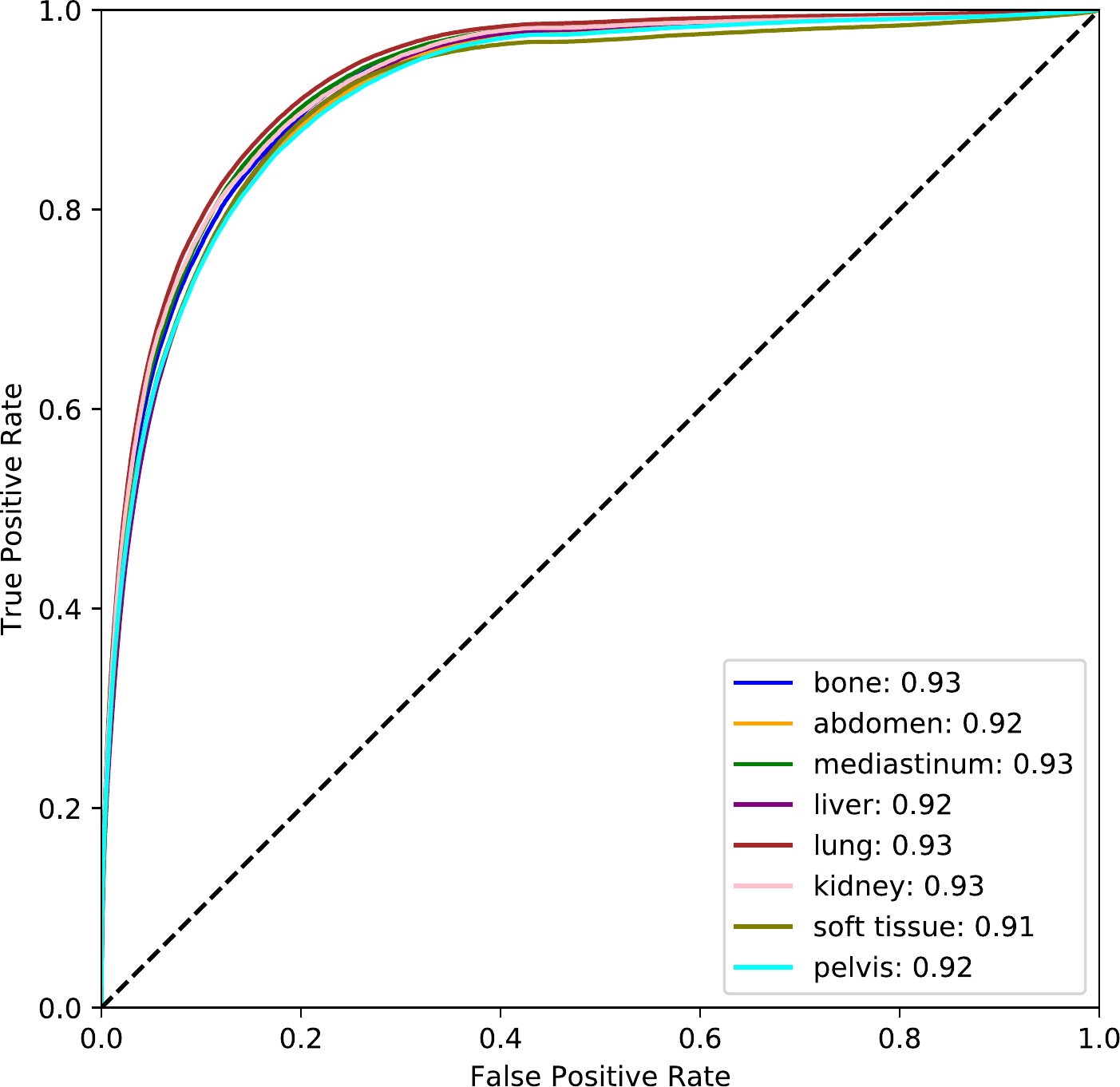}
		\caption{\textit{Supervised by lesion type}}
	\end{subfigure}
	\hspace*{\fill}
	
	\caption{\textit{Pixel-level ROC curves for individual lesion types of DeepLesion data~(\cite{yan2018deeplesion}). FPI and a supervised method are plotted in (a) and (b), respectively.}}
	\label{figures:DeepLesion_ROC_lesionType}
\end{figure*}

For the pixel-level score, only slices with lesions are considered so that we can directly assess localization. The supervised method excels in this setting because the training and test data are consistent. Even so, the supervised DICE score is modest and the others are quite low. This can be partly attributed to the fact that bounding boxes are used as approximate segmentation masks. Although the pixel-level anomaly predictions may not overlap accurately with the complete bounding boxes, the AUROC scores indicate that these regions tend to be rated as more anomalous. This level of performance is insufficient for lesion segmentation, but may be reasonable for highlighting suspicious regions in an anomaly setting. All unsupervised methods achieve an AUROC over 0.5 with FPI scoring the highest among the unsupervised methods. Full ROC curves are plotted in Figure~\ref{figures:DeepLesion_ROC_method} (b). Individual ROC curves for each lesion type are also shown for FPI in Figure~\ref{figures:DeepLesion_ROC_lesionType} (a) and for the supervised method in Figure~\ref{figures:DeepLesion_ROC_lesionType} (b). FPI performs similarly on each lesion type, indicating that it is equally sensitive to a broad range of lesions.

\begin{figure*}[h]
	\centering
    \makebox[\linewidth]{\includegraphics[clip,width=\textwidth]{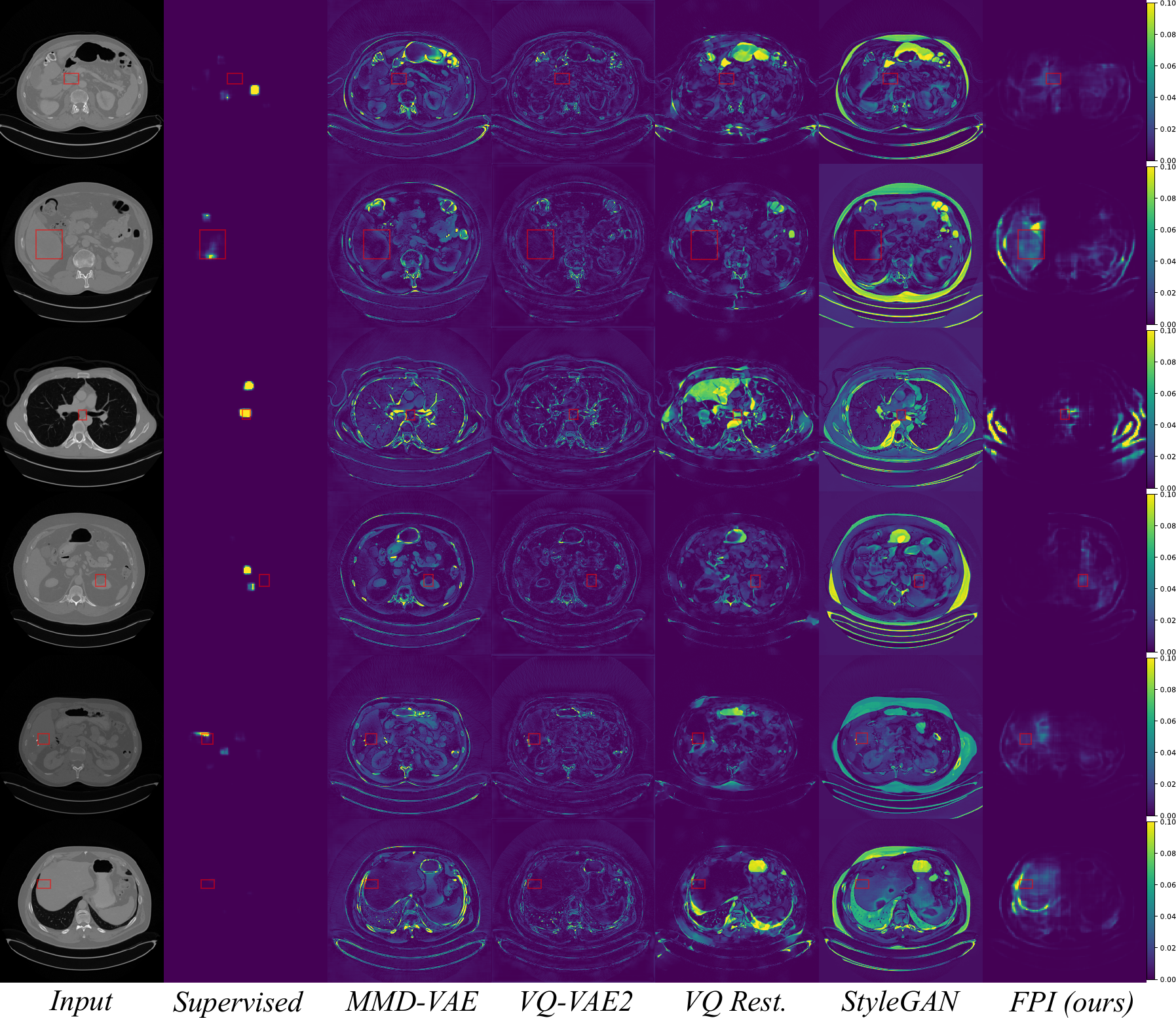}}
	\hspace*{\fill}
	
	\caption{\textit{Anomalous test samples from DeepLesion~(\cite{yan2018deeplesion}) and outputs from each method. Bounding boxes indicate lesions. Note that reconstructiont error outputs are scaled down by a factor of five.}}
	\label{figures:DeepLesion_map_anom}
\end{figure*}

Figure~\ref{figures:DeepLesion_map_anom} displays anomalous examples from the DeepLesion dataset with bounding box labels for each lesion. The outputs from each method show varying levels of sensitivity. MMD-VAE exhibits reconstruction errors throughout the images which reflects the difficulty of learning a compact representation for data with high variation and detail. VQ-VAE2 uses a hierarchical architecture to produce higher fidelity reconstructions with less error. However, this does not help the network to be sensitive to specific irregularities such as lesions. Using the VQ-VAE2 for image restoration can help to highlight regions based on likelihood, rather than purely on intensity differences. This approach can be more selective, but it also tends to highlight certain natural variations that may be deemed less likely. Meanwhile, StyleGAN searches for a normal matching image in its latent space, but it is not always possible to find a good match when the data has complex and detailed structures that can vary greatly across images. In comparison to the reconstruction-based methods, FPI highlights more specific areas in the image that contain lesions or other unusual elements that are not lesions. Finally, the supervised method gives the most lesion-specific activations which can only be learned through labelled examples.

%Supervised is most specific, fpi sensitive to broad range of irregularities, reconstruction models are sensitive to almost every element in the image

\afterpage{\clearpage}
%\clearpage
\section{Discussion}
The proposed method uses a simple self-supervised task to simulate subtle irregularities in the image. Our ablation study suggests that two aspects of this task are important, exposure and difficulty. Without these qualities, the network can overfit to the self-supervised task and fail to detect other types of anomalies. For instance, generating samples with a binary interpolation factor limits the network's exposure to samples with swapped patches. This leads to poor generalization to other types of synthetic anomalies (Figure~\ref{figures:avgPrec}, `Binary'). A varying interpolation factor provides exposure to abnormalities with varying levels of subtlety. However, exposure is not sufficient on its own. The challenge of estimating the \textit{value} of the interpolation factor is also crucial. If training examples are created using a varying interpolation factor ($\alpha \in [0,1]$), but the task is simplified by rounding the label to a binary value ($\alpha = 1 \; \textrm{if} \; \alpha > 0$) then generalization is also poor (Figure~\ref{figures:avgPrec}, `Continuous Round-up'). The difficulty and variety of the proposed task allow FPI to achieve high performance, whether using continuous or discrete $\alpha$ values. Stochastic weight averaging can also provide some benefit, particularly in pixel-wise scores on our synthetic test data (Figure~\ref{figures:avgPrec}). Nonetheless, it is not strictly necessary and good results can be achieved without it. 

Due to the nature of the self-supervised task and the synthesized outliers, one concern is that the network may only detect artifacts, such as discontinuities in image intensity. Indeed, if the characteristics of the synthetic anomalies are more consistent than the characteristics of the normal data, then the network may learn to recognize these artifacts instead of learning the normal appearance of healthy anatomy, which is the real goal. As such, we evaluate FPI using a range of synthetic anomalies, including intensity shifts and deformations; global anomalies that have no discontinuities; and real medical anomalies. The results demonstrate that FPI can detect a broad range of abnormalities, even if there are no discontinuities. This implies that the self-supervised task helps the network to learn the normal appearance of anatomy to some extent. Any deviations from that expectation are therefore seen as foreign patterns being introduced ($\alpha>0$). %Make the abnormality harder to learn than the normal

A major difference between this work and reconstruction-based methods is that we focus on subtle irregularities. In a reconstruction-based approach, the abnormality score is directly proportional to the intensity differences between the test image and its reconstruction. This makes it difficult to detect more subtle irregularities, especially if the normal data has a high variance and is more difficult to faithfully reconstruct. The DeepLesion dataset exhibits both of these characteristics. The lesions can be very subtle and the anatomy varies considerably. In some cases the field of view is centered on the anatomy of interest and other structures are missing or misaligned. Our evaluation on the DeepLesion dataset indicates that reconstruction-based methods are sensitive to gross intensity differences and variations in anatomy. They are largely unable to selectively highlight subtle lesions (Figure~\ref{figures:DeepLesion_map_anom}). Image level AUROC for both reconstruction-based methods is actually below 0.5 (Table~\ref{tab:deepLesionAuroc}). This means that reconstruction error is higher in some normal slices than it is in abnormal slices. This could be because normal slices are peripheral to the lesion slices and may have more variance in structure. This in turn can raise the reconstruction error which is dominated by larger structural differences in the image. In contrast, FPI is able to ignore most variations in normal anatomy. Rather than trying to reconstruct every detail, FPI is trained to detect only regions that are incongruous with the rest of the image (\emph{i.e.},  foreign patches). This allows FPI to be more sensitive to subtle irregularities such as lesions. In this way, FPI can complement reconstruction-based methods and detect less obvious cases that might otherwise require more intense scrutiny. 
%subtle irregularities. These can be difficult and time consuming.

One challenge in unsupervised outlier detection is selecting the best model. Validation sets can be used to select the most performant model. However, this may introduce a bias toward the types of outliers in the validation set. Even if the validation set is disjoint from the test set, there are likely similarities. This may lead to overestimation of performance and failure on unexpected outliers encountered during deployment. As such, we avoid using outliers for validation and simply keep the training duration fixed. Using the same training regime we demonstrate FPI's capability across several datasets. For real world deployment, it may be important to add elements such as uncertainty estimation to make predictions more informative.

\section{Conclusion}
We propose a self-supervision framework for detecting fine-grained abnormalities, common in medical data. Foreign patterns are drawn from independent subjects and used to simulate abnormalities. The network is trained to detect where and to what degree a foreign pattern has been introduced. The resulting model is able to generalize to a wide range of subtle irregularities and achieved the highest rank in the 2020 MICCAI MOOD challenge~(\cite{zimmerer2020medical}) in both sample and pixel level tasks. We also demonstrate FPI's ability to detect a broad range of real medical lesions in the challenging DeepLesion dataset.

The goal of future work is to improve performance on cases where there is less structural consistency. Further extensions could also provide uncertainty estimates for the predicted anomaly scores. Ultimately we hope to reduce the burden placed on radiologists. 

\acks{JT was supported by an Imperial College London President's Scholarship. JB was supported by the UKRI CDT in AI for Healthcare http://ai4health.io (Grant No. P/S023283/1). This work was supported by the London Medical Imaging \& AI Centre for Value Based Healthcare (104691), EP/S013687/1, EP/R005982/1 and Nvidia for the ongoing donations of high-end GPUs.}

% \newpage
\clearpage
\appendix
\section{Foreign Patch Interpolation in Brain Images}
\label{appendix:fpiBrain}
\begin{figure*}[h]
	\centering
	\begin{subfigure}[t]{0.3\linewidth}
        \includegraphics[trim={0cm 6.77cm 0cm 6.77cm},clip,height=0.85\textheight,width=\linewidth,keepaspectratio]{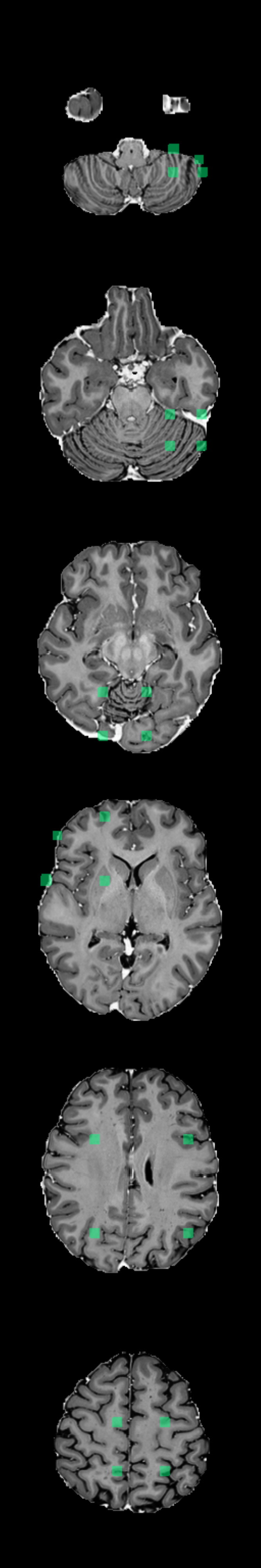}
		\caption{\textit{Original}}
	\end{subfigure}
	\hspace*{\fill}
	\begin{subfigure}[t]{0.3\linewidth}
		\includegraphics[trim={0cm 6.77cm 0cm 6.77cm},clip,height=0.85\textheight,width=\linewidth,keepaspectratio]{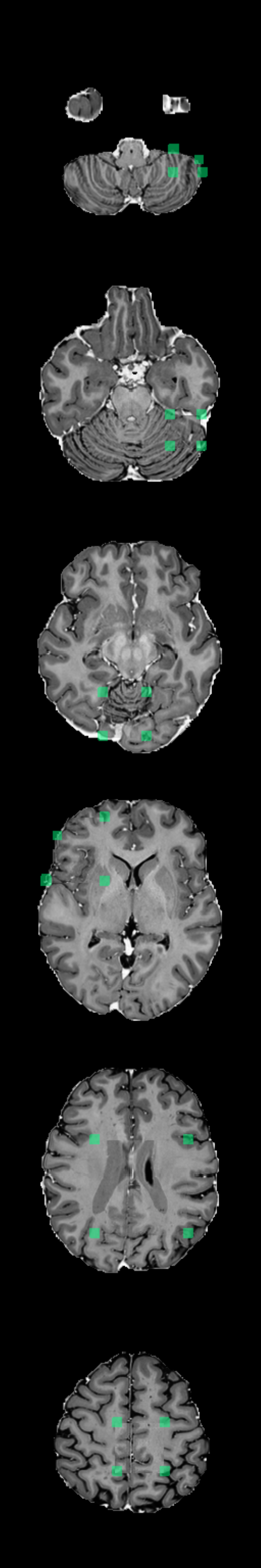}
		\caption{\textit{Interpolation}}
	\end{subfigure}
	\hspace*{\fill}
	\begin{subfigure}[t]{0.3\linewidth}
		\includegraphics[trim={0cm 6.77cm 0cm 6.77cm},clip,height=0.85\textheight,width=\linewidth,keepaspectratio]{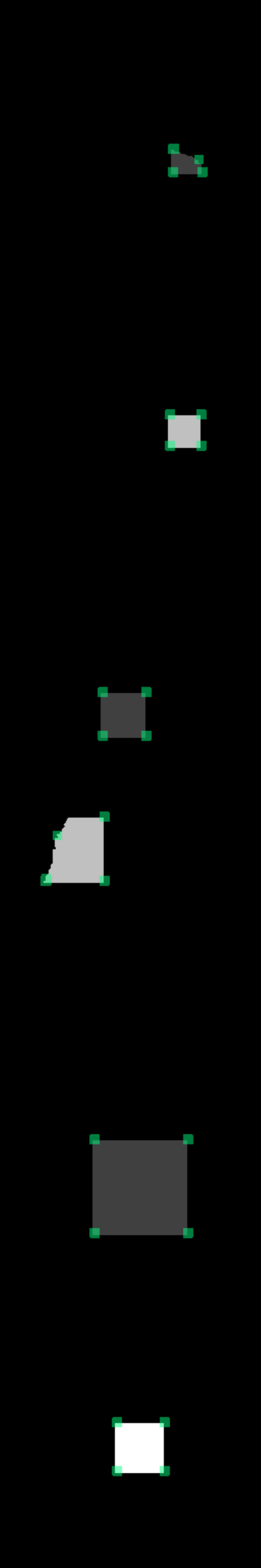}
		\caption{\textit{Label}}
	\end{subfigure}
	\hspace*{\fill}
	\begin{subfigure}[t]{0.05\linewidth}
		\includegraphics[trim={0cm 8.037cm 0cm 8.037cm},clip,height=0.85\textheight,width=\linewidth,keepaspectratio]{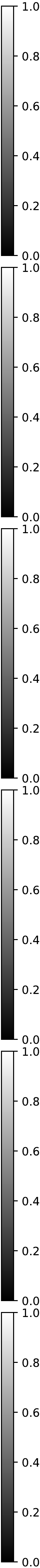}
		%\caption{\textit{Label}}
	\end{subfigure}
	\hspace*{\fill}
	\caption{\textit{MOOD brain images~(\cite{zimmerer2020medical}) with foreign patches.}}
	\label{figures:fpiBrainApp}		
\end{figure*}

\section{Foreign Patch Interpolation in Abdominal Images}
\label{appendix:fpiAbdomen}
\begin{figure*}[h]
	\centering
	\begin{subfigure}[t]{0.3\linewidth}
        \includegraphics[trim={0cm 13.55cm 0cm 13.55cm},clip,height=0.85\textheight,width=\linewidth,keepaspectratio]{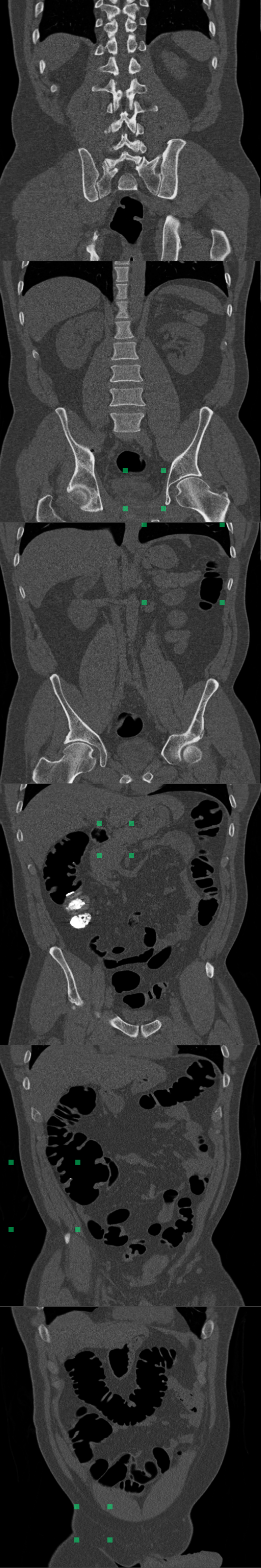}
		\caption{\textit{Original}}
	\end{subfigure}
	\hspace*{\fill}
	\begin{subfigure}[t]{0.3\linewidth}
		\includegraphics[trim={0cm 13.55cm 0cm 13.55cm},clip,height=0.85\textheight,width=\linewidth,keepaspectratio]{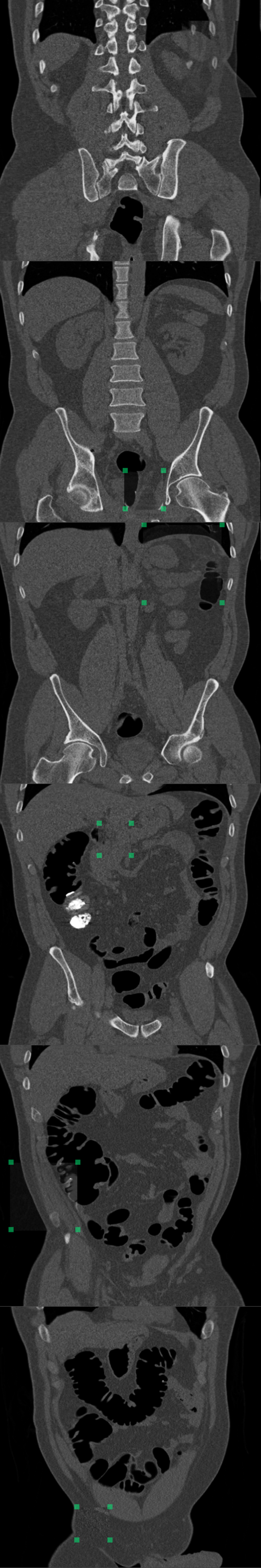}
		\caption{\textit{Interpolation}}
	\end{subfigure}
	\hspace*{\fill}
	\begin{subfigure}[t]{0.3\linewidth}
		\includegraphics[trim={0cm 13.55cm 0cm 13.55cm},clip,height=0.85\textheight,width=\linewidth,keepaspectratio]{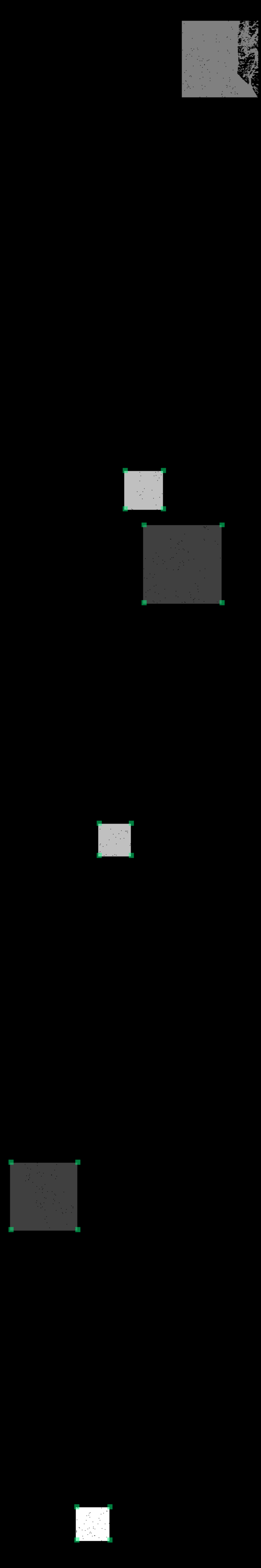}
		\caption{\textit{Label}}
	\end{subfigure}
	\hspace*{\fill}
	\begin{subfigure}[t]{0.05\linewidth}
		\includegraphics[trim={0cm 8.037cm 0cm 8.037cm},clip,height=0.85\textheight,width=\linewidth,keepaspectratio]{selfSupervisedOutlier/figures/fpi_ex_corners/graycolorbarTile.pdf}
		%\caption{\textit{Label}}
	\end{subfigure}
	\hspace*{\fill}
	\caption{\textit{MOOD abdominal images~(\cite{zimmerer2020medical}) with foreign patches.}}
	\label{figures:fpiAbdomApp}		
\end{figure*}

\section{Examples of Synthetic Outliers}
\label{appendix:synthetic}
\begin{figure*}[h]
	\centering
	
	%uniform add
	\begin{subfigure}[t]{0.29\linewidth}
        \includegraphics[trim={0.323cm 0.248cm 0.101cm 0.102cm},clip,height=0.16\textheight,width=\linewidth,keepaspectratio,angle=90]{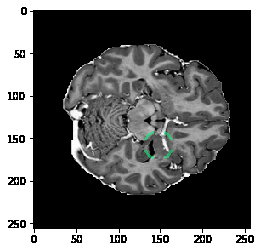}
		%\caption{\textit{Original}}
	\end{subfigure}
	\hspace*{\fill}
	\begin{subfigure}[t]{0.29\linewidth}
		\includegraphics[trim={0.323cm 0.248cm 0.101cm 0.102cm},clip,height=0.16\textheight,width=\linewidth,keepaspectratio,angle=90]{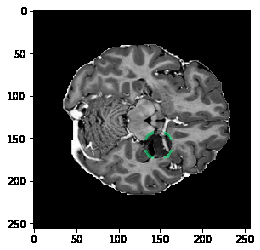}
		%\caption{\textit{Interpolation}}
	\end{subfigure}
	\hspace*{\fill}
	\begin{subfigure}[t]{0.29\linewidth}
		\includegraphics[trim={0.323cm 0.248cm 0.101cm 0.102cm},clip,height=0.16\textheight,width=\linewidth,keepaspectratio,angle=90]{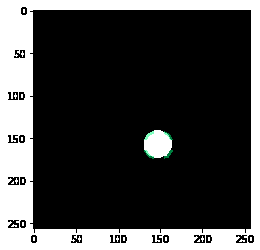}
		%\caption{\textit{Label}}
	\end{subfigure}
	\hspace*{\fill}
	
	%noise add
	\begin{subfigure}[t]{0.29\linewidth}
        \includegraphics[trim={0.323cm 0.248cm 0.101cm 0.102cm},clip,height=0.16\textheight,width=\linewidth,keepaspectratio,angle=90]{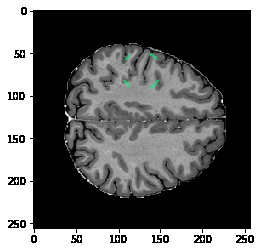}
		%\caption{\textit{Original}}
	\end{subfigure}
	\hspace*{\fill}
	\begin{subfigure}[t]{0.29\linewidth}
		\includegraphics[trim={0.323cm 0.248cm 0.101cm 0.102cm},clip,height=0.16\textheight,width=\linewidth,keepaspectratio,angle=90]{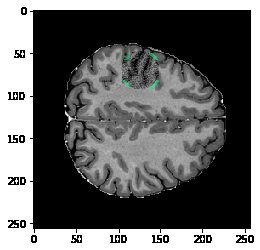}
		%\caption{\textit{Interpolation}}
	\end{subfigure}
	\hspace*{\fill}
	\begin{subfigure}[t]{0.29\linewidth}
		\includegraphics[trim={0.323cm 0.248cm 0.101cm 0.102cm},clip,height=0.16\textheight,width=\linewidth,keepaspectratio,angle=90]{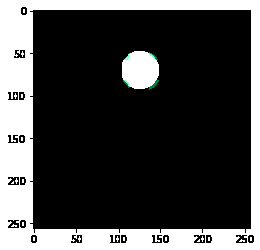}
		%\caption{\textit{Label}}
	\end{subfigure}
	\hspace*{\fill}
	
	%sink/source deform
	\begin{subfigure}[t]{0.29\linewidth}
        \includegraphics[trim={0.323cm 0.248cm 0.101cm 0.102cm},clip,height=0.16\textheight,width=\linewidth,keepaspectratio,angle=90]{selfSupervisedOutlier/figures/testset_corners/ssdeformOrig_corners.pdf}
		%\caption{\textit{Original}}
	\end{subfigure}
	\hspace*{\fill}
	\begin{subfigure}[t]{0.29\linewidth}
		\includegraphics[trim={0.323cm 0.248cm 0.101cm 0.102cm},clip,height=0.16\textheight,width=\linewidth,keepaspectratio,angle=90]{selfSupervisedOutlier/figures/testset_corners/ssdeformApply_corners.pdf}
		%\caption{\textit{Interpolation}}
	\end{subfigure}
	\hspace*{\fill}
	\begin{subfigure}[t]{0.29\linewidth}
		\includegraphics[trim={0.323cm 0.248cm 0.101cm 0.102cm},clip,height=0.16\textheight,width=\linewidth,keepaspectratio,angle=90]{selfSupervisedOutlier/figures/testset_corners/ssdeformLabel_corners.pdf}
		%\caption{\textit{Label}}
	\end{subfigure}
	\hspace*{\fill}
	
	%uniform shift
	\begin{subfigure}[t]{0.29\linewidth}
	    \includegraphics[trim={0.323cm 0.248cm 0.101cm 0.102cm},clip,height=0.16\textheight,width=\linewidth,keepaspectratio,angle=90]{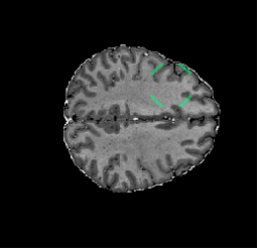}
        %\caption{\textit{Original}}
	\end{subfigure}
	\hspace*{\fill}
	\begin{subfigure}[t]{0.29\linewidth}
	    \includegraphics[trim={0.323cm 0.248cm 0.101cm 0.102cm},clip,height=0.16\textheight,width=\linewidth,keepaspectratio,angle=90]{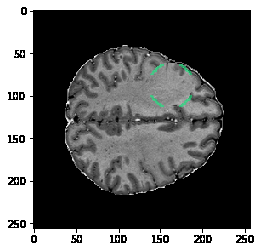}
		%\caption{\textit{Interpolation}}
	\end{subfigure}
	\hspace*{\fill}
	\begin{subfigure}[t]{0.29\linewidth}
	    \includegraphics[trim={0.323cm 0.248cm 0.101cm 0.102cm},clip,height=0.16\textheight,width=\linewidth,keepaspectratio,angle=90]{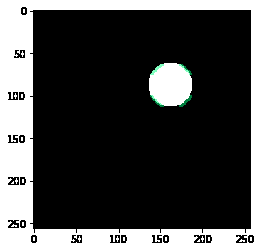}
		%\caption{\textit{Label}}
	\end{subfigure}
	\hspace*{\fill}
	
	%reflect
	\begin{subfigure}[t]{0.29\linewidth}
        \includegraphics[trim={0.323cm 0.248cm 0.101cm 0.102cm},clip,height=0.16\textheight,width=\linewidth,keepaspectratio,angle=90]{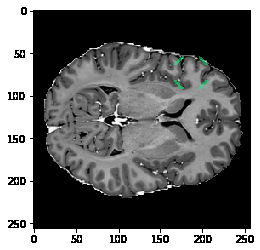}
        \caption{\textit{Original}}
	\end{subfigure}
	\hspace*{\fill}
	\begin{subfigure}[t]{0.29\linewidth}
	    \includegraphics[trim={0.323cm 0.248cm 0.101cm 0.102cm},clip,height=0.16\textheight,width=\linewidth,keepaspectratio,angle=90]{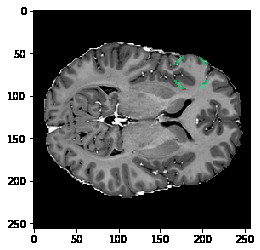}
		\caption{\textit{Synthetic Outlier}}
	\end{subfigure}
	\hspace*{\fill}
	\begin{subfigure}[t]{0.29\linewidth}
		\includegraphics[trim={0.323cm 0.248cm 0.101cm 0.102cm},clip,height=0.16\textheight,width=\linewidth,keepaspectratio,angle=90]{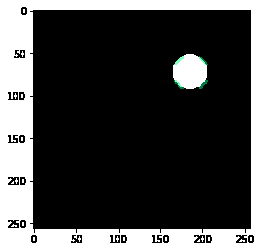}
		\caption{\textit{Label}}
	\end{subfigure}
	\hspace*{\fill}
	\caption{\textit{Each row shows one type of synthetic outlier. From top to bottom these are uniform addition, noise addition, sink/source deformation, uniform shift, and reflection. Original data from MOOD challenge~(\cite{zimmerer2020medical}).}}
	\label{figures:syntheticOutliersApp}		
\end{figure*}

%\noindent

%{\noindent \em Remainder omitted in this sample. }

\clearpage
\vskip 0.2in
\bibliography{ref}

\end{document}